%% file: main.tex
\pdfoutput=1

\documentclass[11pt]{article}

\usepackage[final]{acl}

\usepackage{times}
\usepackage{latexsym}

\usepackage[T1]{fontenc}

\usepackage[utf8]{inputenc}

\usepackage{microtype}

\usepackage{inconsolata}

\input{config}
\input{math_commands}

\title{RulE: Knowledge Graph Reasoning with Rule Embedding}

\author{Xiaojuan Tang,$^{1,3}$ Song-Chun Zhu,$^{1,2,3}$ Yitao Liang,\thanks{Corresponding authors}$^{1,3}$ Muhan Zhang$^{*1,3}$\vspace{1mm}\\
$^1$Institute for Artificial Intelligence, Peking University \hspace{3mm}$^2$Tsinghua University\hspace{3mm}\\$^3$National Key Laboratory of General Artificial Intelligence, BIGAI\\
{\small $^1$ \texttt{{xiaojuan}@stu.pku.edu.cn}} \hspace{3mm} {\small $^1$  \texttt{{\{muhan,yitaol,s.c.zhu\}}@pku.edu.cn}} \\
{\small $^3$\texttt{\{tangxiaojuan,sczhu,liangyitao,mhzhang\}@bigai.ai}}}

\begin{document}
\maketitle

\begin{abstract}
Knowledge graph reasoning is an important problem for knowledge graphs. In this paper, we propose a novel and principled framework called \textbf{RulE} (stands for {Rul}e {E}mbedding) to effectively leverage logical rules to enhance KG reasoning. Unlike knowledge graph embedding methods, RulE learns rule embeddings from existing triplets and first-order {rules} by jointly representing \textbf{entities}, \textbf{relations} and \textbf{logical rules} in a unified embedding space. Based on the learned rule embeddings, a confidence score can be calculated for each rule, reflecting its consistency with the observed triplets. This allows us to perform logical rule inference in a soft way, thus alleviating the brittleness of logic. On the other hand, RulE injects prior logical rule information into the embedding space, enriching and regularizing the entity/relation embeddings. This makes KGE alone perform better too. RulE is conceptually simple and empirically effective. 
We conduct extensive experiments to verify each component of RulE.
Results on multiple benchmarks reveal that our model outperforms the majority of existing embedding-based and rule-based approaches. The code is released at \url{https://github.com/XiaojuanTang/RulE}
\end{abstract}

\section{Introduction}

Knowledge graphs (KGs) usually store millions of real-world facts and are used in a variety of applications~\cite{wang2018ripplenet,bordes2014question,xiong2017explicit}. Examples of knowledge graphs include Freebase \cite{Bollacker2008FreebaseAC}, WordNet \cite{miller1995wordnet} and YAGO \citep{suchanek2007yago}. They represent entities as nodes and relations among entities as edges. Each edge encodes a fact in the form of a triplet ({head entity}, {relation}, {tail entity}). However, KGs are usually highly incomplete, making their downstream tasks more challenging. Knowledge graph reasoning, which predicts missing facts by reasoning on existing facts, has thus become a popular research area in artificial intelligence.

There are two prominent lines of work in this area: \textit{knowledge graph embedding (KGE)} and \textit{rule-based KG reasoning}. Knowledge graph embedding (KGE) methods such as TransE~\cite{bordes2013translating}, RotatE~\cite{sun2019rotate} and BoxE~\cite{abboud2020boxe} embed entities and relations into a latent space and compute the score for each triplet to quantify its plausibility. KGE is efficient and robust to noise. However, it only uses zeroth-order (propositional) logic to encode existing facts (e.g., ``Alice is Bob's wife.'') without explicitly leveraging first-order (predicate) logic. First-order logic uses the universal quantifier to represent \textbf{generally applicable logical rules}. For instance, ``$\forall x, y \colon x ~\text{is} ~ y\text{'s wife} \rightarrow y ~\text{is}~x\text{'s husband}$". Those rules are \textbf{not specific to particular entities} (e.g., Alice and Bob) but are generally applicable to all entities. The other line of work, rule-based KG reasoning, in contrast, explicitly applies logic rules to infer new facts~\cite{galarraga2013amie,galarraga2015fast,yi2018neural,sadeghian2019drum,qu2020rnnlogic}. Unlike KGE, logical rules can achieve interpretable reasoning and generalize to new entities. However, the brittleness of logical rules greatly harms prediction performance. Consider the logical rule $(x,\text{works in}, y) \rightarrow (x, ~\text{lives in}, y)$ as an example. It is mostly correct. Yet, if somebody works in New York but actually lives in New Jersey, the rule can still only infer the wrong fact in an absolute way.

Considering that the aforementioned two lines of work can complement each other, addressing each other's weaknesses with their own merits, it becomes imperative to study how to integrate logical rules with KGE methods in a principled manner. If we view this integration in a broader context, embedding-based reasoning can be seen as a neural method, while rule-based reasoning can be seen as a symbolic method. Neural-symbolic learning has also been a focus of artificial intelligence research in recent years~\cite{parisotto2017neuro,yi2018neural,manhaeve2018deepproblog,xu2018semantic,hitzler2022neuro}.

\setlength{\intextsep}{0pt plus 3pt}
\begin{figure}
\vspace{-0.2cm}
  \centering
  \includegraphics[width=1.0\columnwidth]{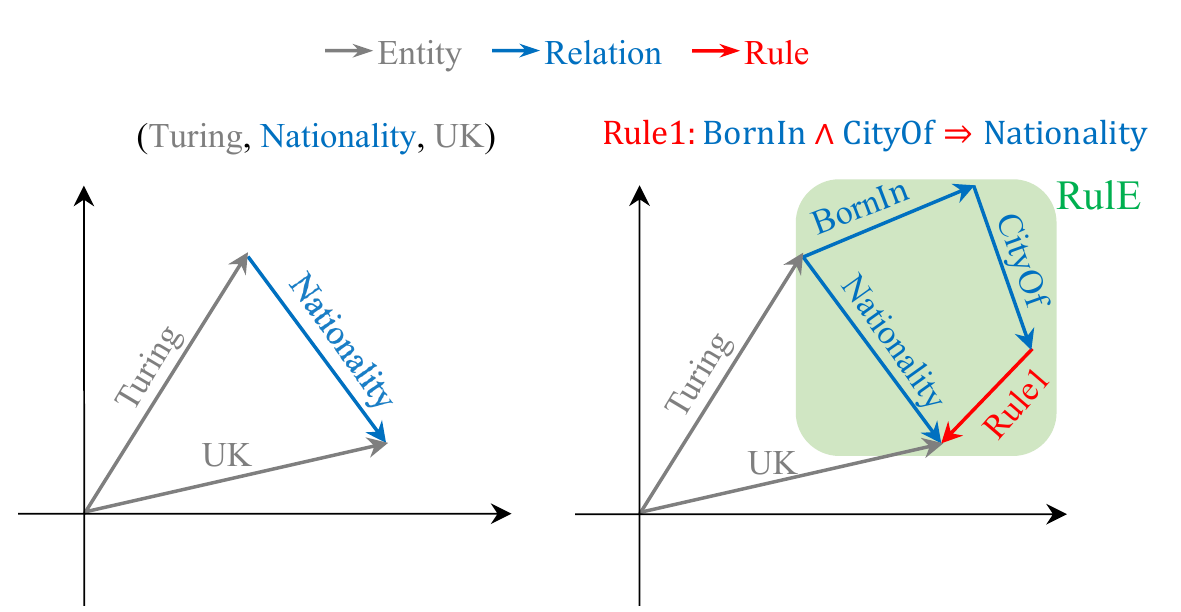} 
\subfloat{(a) Traditional KGE}
\label{fig-1a} 
\quad  \quad   \quad  \quad 
\subfloat{(b) Our RulE}
\label{fig-1b} \quad  
\vspace{-0.2cm}
\caption{(a) Traditional KGE methods embed entities and relations as low-dimensional vectors only using existing triplets by defining operations between entities and relations (e.g., translation); (b) Our RulE 
associates each rule with an embedding and additionally defines mathematical operations between relations and logical rules (e.g., multi-step translation) to leverage first-order \textbf{logical rules}. }
\label{fig:idea}
\end{figure}
   In the KG domain, such efforts exist too. Some works combine logical rules and KGE by using rules to infer new facts as additional training data for KGE~\cite{guo2016jointly,guo2018knowledge} or directly convert some rules into regularization terms for specific KGE models~\cite{ding2018improving,guo2020knowledge}. However, they both leverage logical rules merely to enhance KGE training without actually using logical rules to perform reasoning. In this way, they might lose the important information contained in explicit rules, leading to empirically worse performance than state-of-the-art methods.

   To address the aforementioned limitations, we propose a simple and principled framework called {\em{RulE}}, which aims to learn rule embeddings by jointly representing entities, relations and logical rules in a unified space. As illustrated in  Figure~\ref{fig:idea}, given a KG and logical rules, RulE assigns an embedding to each entity, relation and rule, and defines respective mathematical operators between entities and relations (traditional KGE part) as well as between relations and rules (RulE part). It is important to note that we cannot define operators between entities and rules because rules are not specific to particular entities. By jointly optimizing entity, relation and rule embeddings in the same space, RulE allows injecting prior logical rule information to enrich and regularize the embedding space. Our experiments reveal that this joint embedding can boost KGE methods themselves. Additionally, based on the relation and rule embeddings, RulE is able to give a confidence score to each rule, similar to how KGE gives each triplet a confidence score. This confidence score reflects how consistent a rule is with the existing facts, and enables performing logical rule inference in a soft way by softly controlling the contribution of each rule, which alleviates the brittleness of logic.

We evaluate RulE on benchmark link prediction tasks and show superior performance. Experimental results reveal that our model outperforms the majority of existing embedding-based and rule-based methods. We also conduct extensive ablation studies to demonstrate the effectiveness of each component of RulE. All the empirical results verify that RulE is a simple and effective framework for neural-symbolic KG reasoning.

\section{Preliminaries}
\vspace{-0.1cm}

 A KG consists of a set of triplets $ \mathcal{K} = \{(\text{h},\text{r},\text{t})~|~ \text{h},\text{t} \in \mathcal{E}, \text{r} \in \mathcal{R} \} \subseteq \mathcal{E} \times \mathcal{R} \times \mathcal{E}$, where $\mathcal{E}$ denotes the set of entities and $\mathcal{R}$ the set of relations. For a testing triplet $(\text{h},\text{r},\text{t})$, we define a query as $q=(\text{h},\text{r},\text{?})$. The knowledge graph reasoning (link prediction) task is to infer the missing entity $\text{t}$ based on the existing facts and rules. 

\vspace{-0.2cm}
\subsection{Embedding-based reasoning}
\vspace{-0.1cm}
Knowledge graph embedding (KGE) represents entities and relations as $embeddings$ in a continuous space. It calculates a score for each triplet based on these embeddings via a scoring function. The embeddings are trained so that facts observed in the KG have higher scores than those not observed. The learning goal here is to maximize the scores of positive facts (existing triplets) and minimize those of sampled negative samples.

 \textbf{RotatE}~\cite{sun2019rotate} is a representative KGE method with competitive performance on common benchmark datasets. It maps entities in a complex space and defines relations as element-wise rotations in each two-dimensional complex plane. Each entity and each relation is associated with a complex vector, i.e., $\bm h,\bm r,\bm t \in \mathbb{C}^{k}$, where the modulus of each element in $\bm r$ is fixed to 1 (multiplying a complex number with a unitary complex number is equivalent to a 2D rotation). If a triplet $(\text{h},\text{r},\text{t})$ holds, it is expected that $ \bm t \approx \bm h \circ \bm r $ in the complex space, where $\circ$ denotes the Hadamard (element-wise) product. Formally, the distance function of RotatE is defined as:
 \vspace{-0.2cm}
  \begin{equation} 
      \label{rotate-distance}
     d(\bm h,\bm r,\bm t) =  \parallel \bm h \circ \bm r - \bm t \parallel.
 \end{equation}
  \vspace{-0.2cm}


 \vspace{-0.2cm}
 \subsection{Rule-based reasoning}
 \vspace{-0.2cm}
 Logical rules are usually expressed as first-order logic formulae, e.g., $\forall x,y,z \colon (x,\text r_1,y) \land (y,\text r_2,z) \rightarrow (x,\text r_3,z)$, or $ \text r_1(x,y) \land \text r_2(y,z) \rightarrow \text r_3(x,z)$ for brevity. The left-hand side of the implication ``$\rightarrow$'' is called \textit{rule body} or premise, and the right-hand side is \textit{rule head} or conclusion. 
 Logical rules are often restricted to be closed, which form chains. For a chain rule, successive relations share intermediate entities (e.g., $y$), and the rule head's and rule body's head/tail entity are the same. Chain rules include common logical rules in KG such as symmetry, inversion, composition, hierarchy, and intersection rules. These rules play an important role in KG reasoning. The length of a rule is the number of atoms (relations) that exist in its rule body. A $grounding$ of a rule is obtained by substituting all variables $x,y,z$ with specific entities. If all triplets in the body of a grounding rule exist in the KG, we get a $support$ of this rule. Those rules that have nonzero support are called \textit{activated} rules. When inferring a query $(\text{h},\text{r},?)$, rule-based reasoning enumerates relation paths between head $\text{h}$ and each candidate tail, and uses activated rules to infer the answer. See Appendix~\ref{app:example} for illustrative examples.


\vspace{-0.2cm}
\section{Method}
\vspace{-0.1cm}

\begin{figure*}[t]
\vspace{-1cm}
\begin{center}

\includegraphics[width=0.95\linewidth]{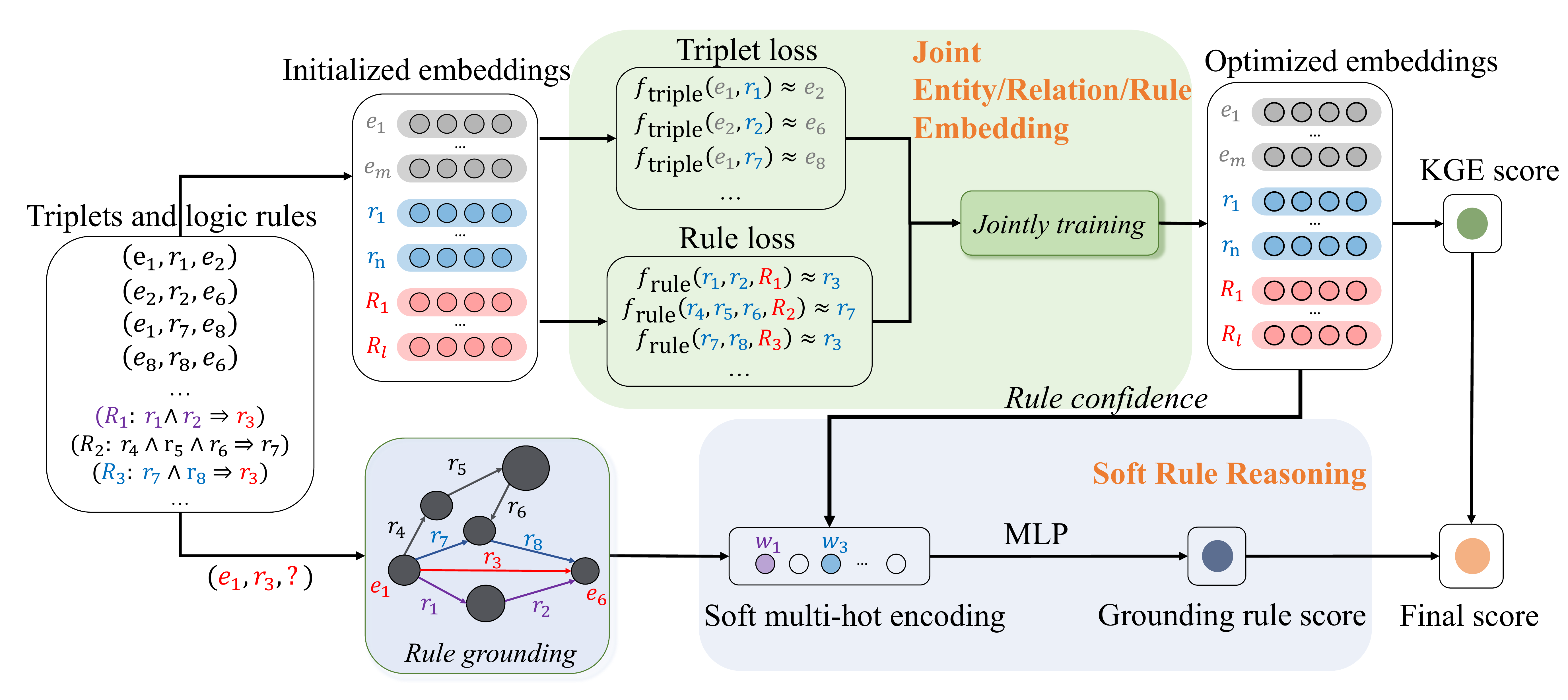}

\end{center}

\caption{Architecture of RulE. It consists of three components. 
1) We first model the relationship between entities and relations as well as the relationship between relations and logical rules to learn \textbf{joint entity, relation and rule embedding} in the same continuous space. With the learned rule embeddings ($\bm R$) and relation embeddings ($\bm r$), RulE can output a weight ($w$) as the confidence score of each rule. 2) In the \textbf{soft rule reasoning} stage, we construct a soft multi-hot encoding $\bm v$ based on rule confidences. Specifically, for triplet $(e_1, r_3, e_6)$, only $R_1$ and $R_3$ can infer the  fact with the grounding paths $e_1 \rightarrow r_1 \rightarrow r_2 \rightarrow e_6$ and $e_1 \rightarrow r_7 \rightarrow r_8 \rightarrow e_6$ (highlighted with purple and blue). Thus, the value of $\bm v_1$ is $w_1$, $\bm v_3$ is $w_3$ and others (unactivated rules) are $0$. Then the constructed soft multi-hot encoding passes an MLP to output the grounding rule score.
3) Finally, RulE \textbf{integrates} the KGE score calculated from the entity and relation embeddings trained in the first stage and the grounding rule score obtained in the second stage to reason unknown triplets.}
\label{fig:overall}

\end{figure*}

This section introduces our proposed model RulE. RulE is a principled framework to combine KG embedding with logical rules by learning rule embeddings. As illustrated in Figure~\ref{fig:overall}, the training process of RulE consists of three key components. Consider a KG containing triplets and a set of logical rules automatically extracted or predefined by experts. They are: 1) \textbf{Joint entity/relation/rule embedding}. We model the relationship between entities and relations as well as the relationship between relations and logical rules to jointly train entity, relation and rule embeddings in a continuous space, as demonstrated in Figure~\ref{fig:idea}. 2) \textbf{Soft rule reasoning}. With the rule and relation embeddings, we calculate a confidence score for each rule which is used as the weight of activated rules to output a grounding rule score. 3) Finally, we \textbf{integrate} the KGE score calculated from the entity and relation embeddings trained in the first stage and the grounding rule score obtained in the second stage to reason unknown triplets. 

\subsection{Joint entity/relation/rule embedding}\label{sec:joint}

Given a triplet $(\text{h},\text{r},\text{t})\in \mathcal{K}$ and a rule $\text{R} \in \mathcal{L}$, we use $\bm h, \bm r,\bm t, \bm R \in \mathbb{C}^{k}$ to represent their embeddings, respectively, where $k$ is the dimension of the complex space (following RotatE). Similar to KGE, which encodes the plausibility of each triplet with a scoring function, RulE additionally defines a scoring function for logical rules. Based on the two scoring functions, it jointly learns entity, relation and rule embeddings in the same space by maximizing the plausibility of existing triplets $ \mathcal{K} $ (zeroth-order logic) and logical rules $\mathcal{L}$ (first-order logic). The following describes in detail how to model the triplets and logical rules together.  

\textbf{Modeling the relationship between entities and relations~}
To model triplets, we take RotatE~\cite{sun2019rotate} due to its simplicity and competitive performance. Its loss function with negative sampling is defined as: 
\begin{equation}
\begin{split}\label{e1}
    L_t{(\bm h,\bm r,\bm t)} = -\log\sigma(\gamma_t - d(\bm h,\bm r,\bm t)) - \\
    ~~~~\sum_{(\bm h^\prime, \bm r, \bm t^\prime) \in \mathbb N}\frac{1}{|\mathbb N|}\log\sigma( d(\bm h,\bm r,\bm t)-\gamma_t ),
\end{split}
\end{equation}
where $\gamma_t$ is a fixed triplet margin, $d(\bm h,\bm r,\bm t)$ is the distance function defined in Equation~(\ref{rotate-distance}), and $\mathbb N$ is the set of negative samples constructed by replacing either the head entity or the tail entity with a random entity using a self-adversarial negative sampling approach. Note that RulE is not restricted to particular KGE models. The RotatE can be replaced with other models, such as TransE~\cite{bordes2013translating} and ComplEx~\cite{trouillon2016complex}, too.

\textbf{Modeling the relationship between relations and logical rules~}
A universal first-order logical rule is some rule that universally holds for all entities. Therefore, we cannot relate such a rule to specific entities. Instead, it is a higher-level concept related only to the relations it is composed of. 
Our modeling strategy is as follows. For a logical rule $\text R \colon \text r_1 \land \text r_2 \land \ldots \land \text r_l \rightarrow \text r_{l+1} $, we expect that $\bm r_{l+1} \approx  (\bm r_1 \circ \bm r_2 \circ \ldots \circ  \bm r_l) \circ \bm R$. Because the modulus of each element in $\bm r$ is restricted to 1, the multiple rotations in the complex plane are equivalent to the summation of the corresponding angles. We define $g(\bm r)$ to return the angle vector of relation $\bm r$ (taking the angle for each element of $\bm r$). Note that the definition of Hadamard product in Equation~\ref{rotate-distance} is equivalent to the term $g (\bm r)$ as defined in Equation~\ref{rule-distance}. More interpretations are provided in Appendix~\ref{app:representations}. Then, the distance function is formulated as follows:
\vspace{-12pt}
\begin{equation}\label{rule-distance}
\begin{aligned}
d_r(\bm r_1,\ldots,\bm r_{l+1}, \bm R) = & \parallel \sum_{i=1}^l  g(\bm r_i) \\
& + g(\bm R) -  g(\bm r_{l+1}) \parallel.
\end{aligned}
\end{equation}

  We also employ negative sampling, the same as when modeling triplets. At this time, it replaces a relation (either in rule body or rule head) with a random relation. The loss function for logical rules is defined as:
\begin{equation}
\begin{aligned}
   & L_r{(\bm r_1,\ldots,\bm r_{l+1}, \bm R)} 
     =  -\log \sigma(\gamma_r-d_r) \\
    & - \sum_{(\bm r_1^\prime,\ldots,\bm r_{l+1}^\prime, \bm R) \in \mathbb M}\frac{1}{|\mathbb M|}\log\sigma(d_r^\prime-\gamma_r),
\end{aligned}
\label{eq:rule-loss}
\end{equation}
where $\gamma_r$ is a fixed rule margin and $\mathbb{M}$ is the set of negative rule samples. 

Note that the above strategy is not the only possible way. For example, when considering the relation order of logical rules (e.g., sister's mother is different from mother's sister), we design a variant of RulE using position-aware sum, which shows slightly improved performance on some datasets. See Appendix~\ref{app:order}. Nevertheless, we find that Equation~(\ref{rule-distance}) is simple and good enough, thus keep it as the default choice.

\textbf{Joint training~}
Given a KG containing triplets $ \mathcal{K} $ and logical rules $\mathcal{L}$, we jointly optimize the two loss functions (\ref{e1}) and (\ref{eq:rule-loss}) to get the final entity, relation and rule embeddings:
\begin{equation}\label{lossall}
\begin{split}
     L &= \sum_{(\bm h,\bm r,\bm t) \in \mathcal{K}}L_t{(\bm h,\bm r,\bm t)} \\ & + \alpha \sum_{(\bm r_1,\ldots,\bm r_l,\bm R) \in \mathcal{L}} L_r{(\bm r_1,\ldots,\bm r_{l+1},\bm R)},
\end{split}
\end{equation}
where $\alpha$ is a hyperparameter to balance the two losses. Note that the two losses act as each other's regularization terms. The rule loss (\ref{eq:rule-loss}) cannot be optimized alone, otherwise there always exist $(\bm r_1,\ldots,\bm r_{l+1},\bm R)$s that can perfectly minimize the loss, leading to meaningless embeddings. However, when jointly optimizing it with the triplet loss, the embeddings will be regularized, and rules more consistent with the triplets tend to have lower losses (by being more easily optimized). On the other hand, the rule loss also provides a regularization to the triplet (KGE) loss by adding additional constraints that relations should satisfy. This additional information enhances the KGE training, leading to entity/relation embeddings more consistent with prior rules. 

\subsection{Soft rule reasoning}\label{sec:soft-reasoning}
As shown in Figure~\ref{fig:overall}, during soft rule reasoning, we use the joint relation and rule embeddings to compute the confidence score of each rule. Similar to how KGE gives a triplet score, the confidence score of a logical rule $\text R_i \colon \text r_{i_1} \land \text r_{i_2} \land...\land \text r_{i_l} \rightarrow \text r_{i_{l+1}}$ is calculated by:
\begin{equation}
    w_i = \gamma_r -  d( \bm r_{i_1},\ldots,\bm r_{i_{l+1}},\bm R_i),
\label{eq:rule_confidence}
\end{equation}
where $d( \bm r_{i_1},\ldots,\bm r_{i_l+1},\bm R_i)$ is defined in Equation~(\ref{rule-distance}). 


To predict a triplet, we perform rule grounding by finding all paths connecting the head and tail that can activate some rule. Often a triplet can have several different rules activated, each with different number of supports (activated paths). An example is shown in Figure~\ref{fig:overall}. The triplet $(e_1, r_3, e_6)$ can be predicted by rule $R_1$ and $R_3$ with the grounding paths $e_1 \rightarrow r_1 \rightarrow r_2 \rightarrow e_6$ and $e_1 \rightarrow r_7 \rightarrow r_8 \rightarrow e_6$. In this case, a straightforward way is to use the maximum (i.e., $\max(w_1, w_3)$) or summation (i.e., $w_1 + w_3$) of the confidences of those activated rules as the grounding rule score of the triplet. 

However, the above way will lose the \textit{dependency} among different rules. For example, consider the following two rules: $\text{parent\_of} (x,y) \rightarrow \text{mother\_of}(x,y)$ and $\text{sister\_of}(x,z) \land \text{aunt\_of}(z,y) \rightarrow \text{mother\_of} (x,y)$. We know that they individually are both not reliable, because a parent can also be a father, and an aunt's sister can be another aunt. However, when these two rules are activated together, one can almost surely infer the ``mother'' relation. In practice, those rules extracted automatically may contain a lot of redundancy or noise. Compared to the naive aggregation approach (such as summation or maximum), we choose to use an MLP to model the \textbf{complex interdependencies} among rules.


Specifically, let us still consider the example in Figure~\ref{fig:overall}. We construct a soft multi-hot encoding $\bm v \in \mathbb R^{|\mathcal{L}|}$ such that $\bm v_i$ is the product of the confidence of $\text R_i$ and the number of grounding paths activating $\text R_i$ (\# of supports). Formally, $\bm v_i = w_i \times |\mathcal{P}(\text h, \text r, \text t, \text R_i)| $ for $i \in \{1,\ldots,\mathcal{L}\}$, where $\mathcal{P}(\text h, \text r, \text t, \text R_i)$ is the set of supports of the rule $R_i$ applying to the current triplet $(\text h, \text r, \text t)$. For the candidate $e_6$ in Figure~\ref{fig:overall}, the value of $\bm v_1$ is $w_1 \times 1$ (grounding path $e_1 \rightarrow r_7 \rightarrow r_8 \rightarrow e_6 $ appears one times), $\bm v_3$ is $w_3 \times 1$, and others (unactivated rules) are $0$.


\begin{table*}[t]
  \caption{Results of reasoning on FB15k-237, WN18RR and YAGO3-10. H@k is in \%. [*] means the numbers are taken from the original papers\protect\footnotemark. [\protect$^\dagger$] means we rerun the methods with the same evaluation process. Best results are in \textbf{bold} while the seconds are \underline{underlined}.}
  \label{FW}
  \resizebox{0.8\textwidth}{!}{
  \begin{tabular}{c|cccc|cccc|cccc}
    \toprule
    \multicolumn{1}{c}{\multirow{1}{*}{}} & \multicolumn{4}{|c|}{\textbf{FB15k-237}} & \multicolumn{4}{c|}{\textbf{WN18RR}}& \multicolumn{4}{c}{\textbf{YAGO3-10}} \\
    
  &MRR &H@1 &H@3 &H@10 &MRR &H@1 &H@3 &H@10 &MRR &H@1 &H@3 &H@10  \\
   \midrule
$\text{TransE}^{\dagger}$   &0.329  &23.0  &36.9  &52.8  &0.222 &1.2 &39.9 &53.0 &0.501 &40.6 &-  &67.4 \\
$\text{DistMult}^*$ &0.241 &15.5 &26.3 &41.9  &0.43  &39  &44  &49 &0.34  &24 &38  &54 \\
$\text{ComplEx}^*$  &0.247  &15.8  &27.5  &42.8  &0.44  &41  &46 &51 &0.36  &26  & 40  & 55 \\
$\text{ConvE}^*$      &0.325  &23.7  &35.6  &50.1  &0.43  &40  &44  &52  &0.44  & 35  &49  &62\\
$\text{TuckER}^*$  &\underline{0.358}  &\textbf{26.6} &\underline{39.4}  &\underline{54.4}  &0.470  &44.3  &48.2  &52.6  &0.529  &-  &-  &67.0 \\
$ \text{RotatE}^{\dagger}$  &0.337  &23.9  &37.4  &53.2  & 0.476  &43.1  &49.2 
 &56.2  &0.497  &40.3  &55.2  &67.5 \\
\midrule
  $\text{PathRank}^*$ &0.087 &7.4 &9.2  &11.2  &0.189  &17.1 &20.0 &22.5  &-  &-  &-   &-\\
 $\text{Neural-LP}^*$  &0.237 &17.3 &25.9  &36.2  &0.435  & 37.1  &43.4  &56.6  &-  &-   &-   &-\\
 $\text{DRUM}^* $   &0.343  &25.5  &37.8 &51.6  & 0.486  &42.5  &51.3 &58.6  &-   &-   &-   &-\\
  $\text{RNNLogic+}~(\text{w/o emb.})^*$ &0.299  &21.5 &32.8 &46.4 &0.489  &45.3  &50.6  &56.3 &-   &-   &-   &-\\
  $\text{RNNLogic+}~(\text{w/o emb.})^{\dagger}$ &0.330  &24.3   &36.3  &50.2 &0.502  &46.1  &52.2  &58.5 &0.484  &41.0 &53.8  &61.5\\
  NCRL &0.30 &20.9 &- &47.3 &\textbf{0.67} &\textbf{56.3} &- &\textbf{85.0} &0.38 &27.4 &- &53.6\\
  \midrule
  $\text{RNNLogic+}~(\text{with emb.})^*$   &  0.349  & 25.8  & 38.5  & 53.3  & 0.513  & 47.1  & 53.2 & 59.7 &-   &-   &-   &- \\ 
$\text{RNNLogic+}~(\text{with emb.})^{\dagger}$   &{0.356}  &\underline{26.2}   &{39.3}   &{54.6}  & {0.516}  & 46.9 & \underline{53.7} & \underline{60.4} &0.499 &41.4  &55.1  &65.8 \\ 
  $\text{Naive Combination }^{\dagger}$ &{0.350} &\underline{26.2} &{38.7} &{52.8}  &0.512  &46.9  &53.1  &{59.7} &0.484 &41.0  &53.7  &61.4   \\
 \midrule
 RulE (emb with TransE.) & 0.346  &25.1  &38.5  &53.4  &0.242  &6.7 &37.8 &52.6  &0.510  &41.4  &57.3  &68.2\\
 RulE (emb.) &0.338 &24.1 &37.6 &53.3 & 0.484  &44.3  & 49.9  & 56.3  &\underline{0.530}  &\underline{44.2}  &\underline{58.2}  &\underline{69.0}  \\
 RulE (rule.) &0.335 &24.9  &36.9 &50.4  &{0.514} &{47.3} &{53.3} &{59.7} &0.481  &40.9  &53.2  &61.0\\
 
 RulE (emb \& rule.) &\textbf{0.362} &\textbf{26.6}  &\textbf{40.0}  &\textbf{55.3} &\underline{0.519}  &\underline{47.5}  &\textbf{53.8} &\underline{60.5}  &\textbf{0.535}  &\textbf{44.7}  &\textbf{58.8}  &\textbf{69.4} \\
 
 \bottomrule

\end{tabular}}


\end{table*}

\footnotetext{Except for YAGO3-10, DistMult, ComplEx and TuckER results are taken from~\citet{abboud2020boxe}.}

With this soft multi-hot encoding $\bm v$, we apply an MLP on $ \bm v$ to calculate the grounding rule score:
\begin{equation}
s_g(\text h,\text r,\text t) = \MLP(\bm v).
\label{eq:grounding}
\end{equation}
Note that for a query $(\text h,\text r, ?)$, we will iterate over all candidates $t$, and the grounding paths for all candidates can be efficiently computed by running BFS. The complexity analysis is presented in Appendix~\ref{app:complexity}. Once we have the grounding rule score for all candidate answers, we further use a softmax function to compute the probability of the true answer. Finally, we train the MLP by maximizing the log likelihood of the true answers in the training triplets. Fine-grained implementation details are included in Appendix~\ref{prac-grounding}.

\subsection{Inference}
Finally, during inference, we predict any missing fact with a weight-ed sum of the KGE score ($s_{t}=\gamma_t - d(\bm h, \bm r, \bm t)$) and the grounding rule score (Equation (\ref{eq:grounding})): 
\begin{equation}\label{equ:final}
    s(\bm h,\bm r,\bm t) = s_{t}(\bm h,\bm r,\bm t) + \beta \cdot s_g(\text h,\text r,\text t^\prime),
\end{equation}
 where $\beta$ is a hyperparameter balancing the weights of embedding-based and rule-based reasoning.
 
\section{Experiments}
In this section, we empirically evaluate RulE on several benchmark KGs and show superior performance to existing embedding-based, rule-based methods and hybrid approaches that combine both. Additionally, we also conduct extensive ablation experiments to verify the effectiveness of each component of RulE.
Furthermore, we provide theoretical analysis and case studies in Appendix~\ref{case studies} to provide further insights and understanding.


\footnotetext{For UMLS and Kinship, [*] means the numbers are taken from~\citet{qu2020rnnlogic}; [$^\dagger$] means we rerun the methods with the same evaluation process. For Family, Neural-LP and DRUM results are taken from~\citet{sadeghian2019drum} and others from our rerun results.}

\subsection{Experiment settings}

\paragraph{Datasets}
We choose six datasets for evaluation: FB15k-237~\cite{toutanova2015observed}, WN18RR~\cite{dettmers2018convolutional}, YAGO3-10~\cite{mahdisoltani2014yago3}, UMLS, Kinship, and Family~\cite{kok2007statistical}. More details of data split and logical rules used in the experiments are in Appendix~\ref{statis}.

\textbf{Baselines~}
We compare with a comprehensive suite of embedding and rule-based baselines. (1) \textit{Embedding-based models}: we include TransE~\cite{bordes2013translating}, 
DisMult~\cite{yang2014embedding}, ComplEx~\cite{trouillon2016complex}, ConvE~\cite{dettmers2018convolutional}, TuckER~\cite{balavzevic2019tucker} and RotatE~\cite{sun2019rotate}.
(2) \textit{Rule-based models}: we compare with MLN~\cite{richardson2006markov}, PathRank~\cite{lao2010relational}, as well as popular rule learning methods Neural-LP~\cite{yang2017differentiable}, DRUM~\cite{sadeghian2019drum}, RNNLogic+ (\textit{w/o emb.})~\cite{qu2020rnnlogic} and NCRL~\cite{cheng2023neural}. 
(3) \textit{Joint KGE and logical rules}: we also compare with baselines that ensemble embedding-based and rule-based method, including RNNLogic+ (\textit{with emb.})~\cite{qu2020rnnlogic} and Naive Combination~\cite{meilicke2021naive}. See more introduction to RNNLogic+ in Appendix~\ref{rnnlogic}.
(4) For our \textit{RulE}, we present results of embedding-based, rule-based and integrated reasoning. The first variant only uses KGE scores obtained from joint entity/relation/rule embedding to reason unknown triplets, denoted by RulE (\textit{emb.}). 
The second variant only uses the grounding score calculated from soft rule reasoning, denoted by RulE (\textit{rule.}). The last one is the full model combining both, denoted by RulE (\textit{emb \& rule.}). Furthermore, to sufficiently verify the effect of rule embedding on different KGE models, we also experiment with a variant of RulE (\textit{emb.}) using TransE~\cite{bordes2013translating} as the KGE model, denoted by $\textit{emb with TransE.}$. We conduct additional experiments on more datasets to compare RulE with the graph-based method NBFNet~\cite{zhu2021neural} (see Appendix~\ref{app:nbfnet}).
Considering the relation order of logical rules, we also design another variant of RulE using position-aware sum (see Appendix~\ref{app:order}). 


\textbf{Evaluation protocols~}
We follow the setting in RNNLogic~\cite{qu2020rnnlogic} and evaluate models by Mean Reciprocal Rank (MRR) as well as Hits at N (H@N). 
For above baselines, we carefully tune the parameters and achieve better results than reported in RNNLogic. To ensure a fair comparison, in the KGE part of RulE, we use the same parameters as those used in TransE and RotatE without further tuning them and rerun RNNLogic+ with the same logical rules as RulE (See Appendix~\ref{app:protocol}).

\begin{table*}[t]
\vspace{-0.3cm}
  \caption{Results of reasoning on UMLS, Kinship and Family. H@k is in \%. [*] means the numbers are taken from~\protect\citet{qu2020rnnlogic}; [$^\dagger$] means we rerun the methods with the same evaluation process\protect\footnotemark. Best results are in \textbf{bold} while the seconds are \underline{underlined}. }
  \label{UK}
  \begin{center}

  \begin{small}

  \resizebox{0.8\textwidth}{!}{
  \begin{tabular}{c|cccc|cccc|cccc}
    \toprule
    \multicolumn{1}{c}{\multirow{1}{*}{}} & \multicolumn{4}{|c|}{\textbf{UMLS}} & \multicolumn{4}{c}{\textbf{Kinship}} & \multicolumn{4}{|c}{\textbf{family}}  \\
   
  &MRR &H@1 &H@3 &H@10 &MRR &H@1 &H@3 &H@10 &MRR &H@1 &H@3 &H@10 \\
\midrule
  $\text{TransE}^{\dagger}$ & 0.704  & 55.4 &82.6 &92.9  &0.300 &14.3 &35.2  &63.7  &0.813  &67.5  &94.6  &98.5 \\
 $\text{DistMult}^*$  &0.391 &25.6 &44.5 &66.9 &0.354 &18.9 &40.0 &75.5   &0.680 &53.0  &78.7  &96.6\\
 $\text{ComplEx}^*$   &0.411 &27.3 &46.8 &70.0   &0.418 &24.2 &49.9 &81.2 &0.930  &88.3  &97.6  &\underline{99.1}\\
 $\text{TuckER}^*$  &0.732 &62.5 &81.2 &90.9   &0.603 &46.2 &69.8 &86.3  &- &- &- &-\\
$\text{RotatE}^{\dagger}$ &0.802 &69.6 &89.0 & 96.3   &0.672 &53.8 &76.4 &93.5  & 0.914   &85.3  &97.4  &{99.0}\\
 
 \midrule
 ${\text{MLN}}^*$     &0.688  &58.7  &75.5  &86.9   &0.351  &18.9  &40.8  &70.7 &- &- &- &- \\
 ${\text{PathRank}}^*$ &0.197 &14.8 &21.4 &25.2  &0.369 &27.2 &41.6 &67.3 &- &- &- &- \\
 ${\text{Neural-LP}}^*$ &0.483 &33.2 &56.3 &77.5   &0.302  &16.7  &33.9  &59.6 &0.91  &86.0  &96.0  &{99.0} \\
 $\text{DRUM}^* $     &0.548  &35.8  &69.9  &85.4  &0.334  &18.3  &37.8  &67.5 &0.950  &91.0   &98.0  & {99.0}  \\
 $\text{RNNLogic+}~(\text{w/o emb.})^{\dagger}$  & 0.800  & 70.4 & 87.8  & 94.3  & 0.655  & 50.4 & 76.0  & 94.7  & 0.974  &96.3  &98.5  &98.6 \\
 NCRL &0.78 &65.9 &- &95.1 &0.64 &49.0 &- &92.9 &0.91 &85.2 &- &\textbf{99.3} \\
 \midrule
 $\text{RNNLogic+}~(\text{with emb.})^{\dagger}$ &0.847 &76.7 &\underline{91.6} &\underline{96.9}  &0.714  &58.1  &81.8  &\underline{95.4} &\underline{0.980}  &97.1  &\underline{98.9}  &\underline{99.1}\\
 $\text{Naive Combination}^{\dagger}$ & \underline{0.856}  &\underline{78.5}  & 91.3  & 96.3  &\underline{0.728} &\underline{60.3}  &\underline{82.1}  &\textbf{ 95.7}  &{0.979}  &\underline{97.2}  &98.5  &98.6  \\
 \midrule
 RulE (emb with TransE.)  &0.748  &61.9   &85.2  &93.3 &0.347 &20.7  &39.8  &62.3   &0.820  &68.9  &94.6  &98.6 \\
 RulE (emb.) & 0.807  & 70.6  &89.2  & 96.3  &0.675 &53.8  &77.1  & 93.7 &0.945  &91.0  &97.9  &\underline{99.1}  \\
 RulE (rule.) &0.827 &74.9 &88.9 &95.5  &0.673 &52.8 &77.5 &95.0 &0.975 &96.7  &98.5  &98.6 \\
 RulE ({emb $\&$ rule.})  &\textbf{0.867}  &\textbf{79.7}  &\textbf{92.5} &\textbf{97.2}  &\textbf{0.736} &\textbf{61.5} &\textbf{82.4} &\textbf{95.7} &\textbf{0.984}  &\textbf{97.8}  &\textbf{99.0}  &\underline{99.1}\\
 
 \bottomrule
  \end{tabular}}
  \end{small}
  \end{center}
\vspace{-0.2cm}
\end{table*}



\textbf{Hyperparameter settings~}
By default, we use RotatE~\cite{sun2019rotate} as our KGE model.
We search for parameters according to validation set performance.
The ranges of the hyperparameters in the grid search and final adopted values are provided in Appendix~\ref{app:hyper}.




\vspace{-0.2cm}
\subsection{Results}
\vspace{-0.2cm}


\begin{table}[t]
\vspace{-0.3cm}
  \caption{Results of reasoning on FB15k and WN18. H@k is in \%. [$\protect^\dagger$] means we rerun the methods with the same evaluation process.}
  \label{KGE}
  \begin{center}
\begin{small}

  \resizebox{\columnwidth}{!}{
  \begin{tabular}{c|cc|cc}
    \toprule
    \multicolumn{1}{c}{\multirow{2}{*}{}} & \multicolumn{2}{|c|}{\textbf{FB15k}} & \multicolumn{2}{|c}{\textbf{WN18}} \\

  &MRR  &H@10 &MRR  &H@10 \\

\midrule
 $\text{TransE}^\dagger$ &0.730   &86.4 &0.772    &92.2   \\
 RulE ({emb with TransE.}) &\textbf{0.734}  &\textbf{86.9}  &\textbf{0.775}  &\textbf{95.0}\\
 \midrule
 $\text{ComplEx}^\dagger$ &0.766  &88.3  &0.898   &\textbf{95.2} \\

 
 RulE ({emb with ComplEx.}) &\textbf{0.788}   &\textbf{89.6} &\textbf{0.928}    &94.4 \\
 \bottomrule

  \end{tabular}}
      
\end{small}
\end{center}
\vspace{-0.2cm}
\end{table}

The results are shown in Table~\Cref{FW,UK}. We observe that: (1) RulE outperforms both embedding-based and rule-based methods on most datasets, especially on UMLS and Kinship which show significant improvements. This indicates that combining KGE and rule-based methods with rule embedding can take advantage of both and improve the performance of KG reasoning. (2) Compared with loosely composed methods (i.e., RNNLogic+ (\textit{with emb.}) and Naive Combination), RulE (\textit{emb \& rule.}) obtains better results on all datasets, demonstrating that it is more beneficial for KG reasoning to use rule embedding to bridge embedding-based and rule-based approaches than naively combining them. A detailed analysis is as follows.

\begin{table*}[h]
\vspace{-0.2cm}
    
  \caption{Ablation study on soft rule reasoning part of RulE. H@k is in \%.}
  \label{tab:ablation}
  \begin{center}
\begin{small}

  \begin{tabular}{c|cc|cc|cc|cc|cc}
    \toprule
    \multicolumn{1}{c}{\multirow{1}{*}{}} & \multicolumn{2}{|c|}{\textbf{FB15k-237}} & \multicolumn{2}{|c}{\textbf{WN18RR}} & \multicolumn{2}{|c}{\textbf{UMLS}} & \multicolumn{2}{|c}{\textbf{Kinship}} & \multicolumn{2}{|c}{\textbf{Family}}\\
  &MRR  &H@10 &MRR  &H@10  &MRR  &H@10 &MRR  &H@10 &MRR  &H@10\\

  \midrule
  
   standard  &0.335  &50.4  &{0.514}  &{59.7} &0.827  &95.5  &0.673  &95.0 &0.975  &98.6\\

  sum (w/o MLP)  &0.276    &42.9  &0.390    &50.9 &0.587  &82.0 &0.591   &90.0    &0.877  &97.6\\
  max (w/o MLP)  & 0.256   &18.4  &0.294    &23.4   &0.346  &23.1 &0.373   &21.7   &0.748  &94.9 \\
 hard-encoding  &0.330    &50.2      &0.496  &45.4  &0.791    &94.6   &0.643  &94.0    &0.973 &96.2\\
 \bottomrule
  \end{tabular}
\vspace{-0.4cm}
      
\end{small}
\end{center}
\vspace{-0.2cm}

\end{table*}

\textbf{Embedding logical rules helps KGE~}
We first compare RulE (\textit{emb.}) with RotatE. Note that RulE (\textit{emb.}) and RulE (\textit{emb with TransE.}) only add an additional rule embedding loss to the KGE training and still use KGE scores only for prediction. As presented in Table~\ref{FW} and~\ref{UK}, RulE (\textit{emb.}) and RulE (\textit{emb with TransE.}) both achieve comparable or higher performance than the corresponding KGE models, especially for RulE (\textit{emb with TransE.}), which obtains 4.4\% and 4.7\% absolute MRR gain than TransE on UMLS and Kinship. This indicates that by jointly embedding entities/relations/rules into a unified space, RulE can inject logical rule information to enrich and regularize the embedding space and improve the generalization of KGE. This verifies the effectiveness of joint entity/relation/rule embedding. 

We also observe that the improvement of RulE (\textit{emb with TransE.}) is more significant than RulE (\textit{emb.}). The reason is probably that RotatE is expressive enough to capture many relational patterns of KG, thus more complex logical rules may be needed. 
In Table~\ref{KGE}, we further use TransE and ComplEx as the KGE model of RulE and test on FB15k and WN18 datasets. They both obtain superior performance to the corresponding KGE models (see Appendix~\ref{abl:transe}).

Additionally, we find that RulE (\textit{emb with TransE.}) on UMLS and Kinship achieves more improvement than FB15k-237 and WN18RR. The reason is probably that UMLS and Kinship contain more rule-inferrable facts while WN18RR and FB15k-237 consist of more general facts (like the publication year of an album, which is hard to infer via rules). This phenomenon is observed in previous works too~\cite{qu2020rnnlogic}. To verify it, we perform a data analysis in Appendix~\ref{cycle}.

\textbf{Soft rule reasoning outperforms hard rule reasoning~}
We compare RulE (\textit{rule.}) with rule mining methods. Note that we rerun RNNLogic+ with the same rules as RulE for fair comparisons. From Table~\ref{FW} and \ref{UK}, we can observe that RulE (\textit{rule.}) outperforms existing hard rule reasoning baselines except for WN18RR on NCRL. This demonstrates that soft multi-hot encoding over MLP is more powerful than other ways of performing rule inference.

 \textbf{Comparison with other joint reasoning and rule-enhanced KGE models~}
 We also compare with RNNLogic+ (\textit{emb \& rule.}) and Naive Combination, which separately trains embedding-based and rule-based methods and then only loosely ensemble them. Although the final inference of RulE (\textit{emb \& rule.)} is similar to the above methods (weighted sum over KGE score and grounding rule score), RulE uses rule embedding as a bridge to strengthen KGE and rule reasoning process, by injecting rule information to the KGE embedding space and also extracting rule confidence for soft rule reasoning. This demonstrates that the interaction between embedding-based methods and rule-based methods can further enhance each other and the rule embedding serves as the medium. We further study how the hyperparameter $\beta$ balances both of them. See more details in Appendix~\ref{app:beta}. 

\vspace{-0.2cm}
\subsection{Ablation study}
\vspace{-0.2cm}
This section analyzes whether individual components of the RulE design are useful via ablation experiments. As the usefulness of joint entity/relation/rule embedding has been verified extensively by previous experiments, here we focus on validating the soft rule reasoning part. Specifically, we compare the following RulE versions: (1) \textit{standard}, which is the standard RulE (\textit{rule.}) described in Section~\ref{sec:soft-reasoning}; (2) \textit{hard-encoding}, which only uses hard 1/0 to select activated rules instead of the rule confidence obtained from joint relation/rule embeddings. This is to verify that the confidence scores of logical rules, which are learned through jointly embedding KG and logical rules, help rule-based reasoning; (3) \textit{sum (w/o MLP)} and \textit{max (w/o MLP)}, which replace the MLP layer with sum and max respectively over the weights of all activated rules as the grounding rule score. This is to demonstrate the importance of capturing the complex interdependencies among logical rules.


\textbf{Ablation Results~} 
As presented in Table~\ref{tab:ablation},
\textit{standard} achieves better performance than \textit{hard-encoding}, which indicates that
using soft multi-hot encoding to perform logical rule inference in a soft way is beneficial to the rule reasoning process. Besides, the performances of \textit{sum (w/o MLP}) and \textit{max (w/o MLP}) versions degrade sharply compared to \textit{standard}, showing that it is important to use an MLP to capture the complex interdependencies among rules.

\vspace{-0.2cm}
\section{Related work}
\vspace{-0.2cm}
\textbf{Embedding-based methods~}
Embedding-based methods aim to learn embeddings for entities and relations and estimate the plausibility of unobserved triplets based on these learned embeddings~\citep{bordes2013translating,yang2014embedding,trouillon2016complex,sun2019rotate,balavzevic2019tucker,vashishth2019composition,zhang2020interstellar,abboud2020boxe,ge2023compounding}.


\textbf{Rule-based methods~}
Learning logical rules for knowledge graph reasoning has also been extensively studied, including Inductive Logic Programming~\cite{quinlan1990learning}, Markov Logic Networks~\cite{kok2005learning,beltagy2014efficient}, AMIE~\cite{galarraga2013amie}, AMIE+~\cite{galarraga2015fast}, Neural-LP~\cite{yang2017differentiable}, DRUM~\cite{sadeghian2019drum}, RNNLogic~\cite{qu2020rnnlogic} and other methods~\cite{cheng2023neural,nandi2023simple}. They almost solely use the learned logical rules for reasoning, which suffer from brittleness and are hardly competitive with embedding-based reasoning in most benchmarks. 

\textbf{Joint KGE and logical rules~}
Some work tries to incorporate logical rules into KGE models.
They usually use logical rules to infer new facts as additional training data for KGE~\cite{guo2016jointly, guo2018knowledge} or inject rules via regularization terms during training~\cite{wang2015knowledge,ding2018improving}. However, they do not really perform reasoning with logical rules. 

\textbf{GNN-based methods~}
Recently, there are some KG reasoning works based on graph neural networks~\cite{schlichtkrull2018modeling,teru2020inductive,zhang2020efficient,zhu2021neural,li2023message}. They exploit neighboring information via message-passing mechanisms. More details of related work and comparison with RNNLogic~\cite{qu2020rnnlogic} are provided in Appendix~\ref{app:related_work}.



\vspace{-0.2cm}
\section{Conclusion}
\vspace{-0.2cm}
We propose a simple and principled framework RulE to jointly represent entities, relations and logical rules in a unified embedding space. The incorporation of rule embedding allows injecting rule information to enrich and regularize the embedding space, thus improving the generalization of KGE. Besides, we also demonstrate that with the learned rule embedding, RulE can perform rule inference in a soft way and empirically verify that using an MLP can effectively model the complex interdependencies among rules, thus enhancing rule inference.

\clearpage

\section{Limitations}
A limitation of RulE is that, similar to prior works which apply logical rules for inference, RulE's soft rule reasoning part needs to enumerate all paths between entity pairs, making it difficult to scale. Another limitation is that currently we only consider chain rules provided as prior knowledge. In the future, we plan to explore more efficient and effective rule reasoning algorithms and consider more complex rules. Besides, currently, we focus on chain rules provided as prior knowledge, i.e., Horn clause, a disjunctive clause (a disjunction of literals) with at most one positive. We acknowledge the importance of addressing negation operators, for example, $\forall x,y,z: \neg r_1(x,y) \land r_2(y,z) \rightarrow r_3(x,z)$. In future explorations, we may consider leveraging betaE~\cite{ren2020beta}, a probabilistic embedding framework to handle negation operator in complex multi-hop logical reasoning.

\section*{Acknowledgements}

This work is partially supported by the National Key R\&D Program of China (2022ZD0160300), the National Key R\&D Program of China (2021ZD0114702), the National Natural Science Foundation of China (62276003).


\bibliography{custom}

\clearpage

\section{Related work}\label{app:related_work}

\paragraph{Embedding-based methods}
Embedding-based methods aim to learn embeddings for entities and relations and estimate the plausibility of unobserved triplets based on these learned embeddings~\citep{bordes2013translating,yang2014embedding,trouillon2016complex,cai2017kbgan,sun2019rotate,balavzevic2019tucker,vashishth2019composition,zhang2020interstellar,{abboud2020boxe},ge2023compounding}.
Much prior work in this regard views a relation as some operation or mapping function between entities. Most notably, 
TransE~\cite{bordes2013translating} defines a relation as a translation operation between some head entity and tail entity. It is effective in modelling inverse and composition rules. DistMult~\cite{yang2014embedding} uses a bilinear mapping function to model symmetric patterns. RotatE~\cite{sun2019rotate} uses rotation operation in complex space to capture symmetry/antisymmetry, inversion and composition rules. 
CompoundE~\cite{ge2023compounding} leverages translation, rotation, and scaling operations to create relation-dependent compound operations on head and/or tail entities.
BoxE~\cite{abboud2020boxe} models relations as boxes and entities as points to capture symmetry/anti-symmetry, inversion, hierarchy and intersection patterns but not composition rules.
These approaches learn representations solely based on triplets (zeroth-order logic) contained in the given KG. In contrast, our approach is able to embody more complex first-order logical rules in the embedding space by jointly modeling entities, relations and logical rules in a unified framework. 

\paragraph{Rule-based methods}
Learning logical rules for knowledge graph reasoning has also been extensively studied. 
As one of the early efforts,~\citet{quinlan1990learning} uses Inductive Logic Programming (ILP) to derive logical rules (hypothesis) from all the training samples in a KG.
Markov Logic Networks (MLNs) ~\cite{kok2005learning,brocheler2012probabilistic,beltagy2014efficient} define the joint distribution of given variables (observed facts) and hidden variables (missing facts) such that missing facts can be inferred in the probabilistic graphical model. 
AMIE~\cite{galarraga2013amie} and AMIE+~\cite{galarraga2015fast} first enumerate possible rules and then learn a scalar weight for each rule to encode its quality.
Neural-LP~\cite{yang2017differentiable} and DRUM~\cite{sadeghian2019drum} mine rules by simultaneously learning logic rules and their weights based on TensorLog~\cite{cohen2017tensorlog}. RNNLogic~\cite{qu2020rnnlogic} simultaneously trains a rule generator and reasoning predictor to generate high-quality logical rules. ~\citet{nandi2023simple} propose three augmentations aimed at enhancing the rule set's coverage in RNNLogic-based models. NCRL~\cite{cheng2023neural} infers rule head by recursively merging atomic compositions in rule body. Except for RNNLogic, the above methods solely use the learned logical rules for reasoning, which suffer from brittleness and are hardly competitive with embedding-based reasoning in most benchmarks. Although RNNLogic considers the effect of KGE during inference, it 
pretrains KGE \emph{separately} from logical rule learning without jointly modeling KGE and logical rules in the same space. Most existing works focus on mining rules from observed triplets. In contrast, we focus on the setting where rules are already given (either mined from KG or provided as prior knowledge) and the task is to leverage the rules for better inference. Thus, in principle, our framework can be combined with any rule mining model to improve their rule usage. 




\paragraph{Joint KGE and logical rules}
Some recent work tries to incorporate logical rules into KGE models to improve the generalization performance of KGE reasoning. KALE~\cite{guo2016jointly} and RUGE~\cite{guo2018knowledge} use logical rules to infer new facts as additional training data for KGE. Several other works inject rules via regularization terms during training, including~\citet{wang2015knowledge} and \citet{ding2018improving}. These methods leverage logical rules only to enhance KGE training and do not really perform reasoning with logical rules. Although \citet{meilicke2021naive} combines symbolic and embedding-based methods, it only loosely ensembles the rankings generated by embedding-based and symbolic methods. 
In contrast, our method jointly learns entity/relation/rule embeddings in a unified space, which is shown to enhance KGE itself. With the learned rule embedding, RulE can also perform logical rule inference in a soft way, improving the rule-based reasoning process. Moreover, the combination of both further advance the performance.

\paragraph{GNN-based methods}
Recently, there are some KG reasoning works based on graph neural networks~\cite{schlichtkrull2018modeling,teru2020inductive,zhang2020efficient,zhu2021neural,li2023message}. They exploit neighboring information via message-passing mechanisms, which are empirically powerful and can be applied to the inductive setting. However, they usually suffer from high complexity. Furthermore, these methods cannot leverage prior/domain knowledge presented as logical rules, its interpretability is built on path-explanation of the predictions.

\section{Example of rule-based reasoning}\label{app:example}
The length of a rule is the number of atoms (relations) that exist in its rule body. One example of a length-2 rule is:
 \begin{equation}\label{rule}
    \text{born\_in}(x,y) \land \text{city\_of}(y,z) \rightarrow \text{nationality}(x,z),
 \end{equation}
 of which $\text{born\_in}(\cdot) \land \text{city\_of}(\cdot)$ is the rule body and $\text{nationality}(\cdot)$ is the rule head. A $grounding$ of a rule is obtained by substituting all variables $x,y,z$ with specific entities. For example, if we replace $x,y,z$ with Bill Gates, Seattle, US respectively, we get a grounding: 
\begin{equation}\label{grounding}
 \begin{split}
     &\text{born\_in}  ({\text{Bill Gates}},{\text{Seattle}})  \land \text{city\_of}({\text{Seattle}},{\text{US}}) \\
    & ~~~~~~~~~~~~~~~\rightarrow \text{nationality}({\text{Bill Gates}},{\text{US}})
 \end{split}
 \end{equation}
If all triplets in the body of a grounding rule exist in the KG, we get a $support$ of this rule. Those rules that have nonzero support are called \textit{activated} rules. When inferring a query $(\text{h},\text{r},?)$, rule-based reasoning enumerates relation paths between head $\text{h}$ and each candidate tail, and uses activated rules to infer the answer. For example, if we want to infer $\text{nationality} ({\text{Bill Gates}},?)$, given the logical rule~(\ref{rule}) as well as the existing triplets $\text{born\_in} ({\text{Bill Gates}}, {\text{Seattle}})$ and $\text{city\_of} ({\text{Seattle}},{\text{US}})$, the answer $\text{US}$ can be inferred.
 
\section{Fine-grained implementation details}\label{prac-grounding}
This section introduces the fine-grained implementation details. Recall the soft reasoning process: we use the joint relation and rule embeddings to compute a \textit{scalar} as the confidence score of each rule, then construct a soft multi-hot encoding with the confidence, and finally pass the MLP layer to output the grounding rule score. In other words, we obtain the grounding rule score by using a multi-hot encoding vector to activate an MLP. However, in practice, we can use a fine-grained way, i.e., use multiple multi-hot encoding vectors rather than only one. 

Specifically, recall that $\bm{R}, \bm r \in \mathbb{C}^k$ are the embeddings of logical rules and relations, respectively. To prevent confusion, we use $\bm v[i]$ to denote the $i$-th elements of vector $\bm v$. With the optimized relation and rule embeddings, we can compute the confidence vector of a logical rule $\text R_i \colon \text r_{i_1} \land \text r_{i_2} \land ... \land \text r_{i_l} \rightarrow \text r_{i_{l+1}}$ as:
\begin{equation}
     \bm c_i = \frac{\gamma_r}{k} - ( \sum_{j=1}^l \bm r_{i_j} + \bm R_i - \bm r_{i_{l+1}} )^p,
\end{equation}
where $p$ is a hyperparameter, usually the same as the norm defined in Equation~(\ref{rule-distance})           
, $\gamma_r$ is the fixed rule margin defined in Equation~(\ref{eq:rule-loss}). Note that $\bm c_i$ is a k-dimensional vector, slightly different from the definition in Section~\ref{sec:soft-reasoning}. Each element of $\bm c_i$ represents a way of encoding the confidence of rule $\text{R}_i$. Given the confidence vector $\bm c_i$, we can further construct $k$ multi-hot encoding vectors. Each multi-hot encoding vector activates the MLP to output a grounding score. Further, the mean of all the grounding scores is computed as the grounding rule score $s_g$ of a triplet. 

Let us consider the example ($e_1, r_3, e_6$) in Figure~\ref{fig:overall}. We construct $k$ soft multi-hot encoding vectors $\{\bm v_j \in \mathbb R^{|\mathcal{L}|},j=1,\ldots,k\}$ such that $\bm v_j[i]$ is the product of of the confidence of $\text R_i$ and the number of grounding paths activating $\text R_i$. Formally, $\bm v_j[i] = \bm c_i[j] \times |\mathcal{P}(\text h, \text r, \text t, \text R_i)| $ for $i \in \{1,\ldots, \mathcal{L}  \}$, where $\mathcal{P}(\text h, \text r, \text t, \text R_i)$ is the set of supports of the rule $\text R_i$ applying to the current triplet $(\text h, \text r, \text t)$. For the candidate $e_6$ in Figure~\ref{fig:overall}, the value of multi-hot encoding vector $\bm v_j[1]$ is $\bm c_1[j] \times 1$, $\bm v_j[3]$ is $\bm c_3[j] \times 1$, and others are 0 (i.e., $\bm v_j[k] = 0, k=2,4,\ldots,\mathcal{L}$).

With these soft multi-hot encoding vectors, we apply an MLP to output the grounding rule score:

\begin{equation}
    s_g = \frac{1}{k} \sum_{j=1}^k \MLP(\bm v_j).
\end{equation}
Note that the MLP used by different soft multi-hot encodings is the same. Once we have the grounding rule score for all candidate answers, we further use a softmax function to compute the probability of the true answer. Finally, we optimize the MLP and grounding-stage rule embedding by maximizing the log likelihood of the true answers based on these training triplets.

\section{Introduction of RNNLogic+}\label{rnnlogic}
RNNLogic~\cite{qu2020rnnlogic} aims to learn logical rules from knowledge graphs, which simultaneously trains a rule generator as well as a reasoning predictor. The former is used to generate rules while the latter learns the confidence of generated rules. Because RulE is designed to leverage the rules for better inference, to compare with it, we only focus on the reasoning predictor RNNLogic+, which is a more powerful predictor than RNNLogic. The details are described in this section. 

Given a KG containing a set of triplets and logical rules, RNNlogic+ associates each logical rule with a grounding-stage rule embedding $\bm R^{(g)}$ (different from the joint rule embedding in RulE), for a query $(\text h, \text r, \text ?)$, it grounds logical rules into the KG, finding different candidate answers. For each candidate answer $\text t^\prime$, RNNLogic+ aggregates all the rule embeddings of those activated rules, each weighted by the number of paths activating this rule (\# supports). Then an MLP is further used to  project the aggregated embedding to the grounding rule score $s_r(\text h, \text r, \text t^\prime)$:
\begin{equation}
s_r = \MLP\big(\AGG( \{\bm R_i^{(g)},|\mathcal{P}(\text h, \text R_i, \text t^\prime)|\}_{\text R_i \in \mathcal{L}} )\big)
\label{eq:rnn-rule-score}
\end{equation}
where $\LN$ is the layer normalization operation, $\AGG$ is the PNA aggregator~\cite{corso2020principal}, $\mathcal{L}$ is the set of generated high-quality logical rules, and $\mathcal{P}(\text h, \text R_i, \text t^\prime)$ is the set of supports of the rule $\text R_i$ which starts from h and ends at $\text t^\prime$. Once RNNLogic+ computes the score of each candidate answer, it can use a softmax function to compute the probability of the true answer. Finally, the predictor can be optimized by maximizing the log likelihood of the true answers based on training triplets. In essence, when replacing the PNA aggregator with sum aggregation, it is equivalent to using hard multi-hot encoding to activate an MLP (i.e., only using hard 1/0 to select activated rules). However, RulE additionally employs the confidence scores of rules as soft multi-hot encoding.


During inference, there are two variants of models: 
\begin{itemize}

    \item RNNLogic+ (\textit{w/o emb.}): 
    This variant only uses the logical rules for knowledge graph reasoning. Specifically, we calculate the score $s_r$ of each candidate answer defined in Equation~(\ref{eq:rnn-rule-score}).
    
    \item RNNLogic+ (\textit{with emb.}):
    It uses RotatE~\cite{sun2019rotate} to \textit{pretrain} knowledge graph embeddings models, which is different from RulE in that RulE jointly models KGE and logical rules in the same space to learn entity, relation and logical rule embeddings. During inference, it linearly combines the grounding rule score and KGE score as the final prediction score, i.e.,
    \begin{equation}
        s(\text h, \text r, \text t^\prime) = s_r(\text h, \text r, \text t^\prime) + \alpha * \text{KGE} (\text h, \text r, \text t^\prime),
    \end{equation}
    where $\text{KGE} (\text h, \text r, \text t^\prime)$ is the KGE score calculated with entity and relation embeddings optimized by RotatE alone, and $\alpha$ is a positive hyperparameter weighting the importance of the knowledge graph embedding score.
\end{itemize}


\section{Analysis of rule-inferrable indicator}\label{cycle}
This section analyzes the rule-inferrable of KGs. Naturally, without considering the directions of edges, any rule can be viewed as a cycle by including both the relation path and the target relation itself. To simplify the analysis, we assume that any cycle can be a logical rule, regardless of concrete relations and the correct semantic information. If a relation appears in a rule, it must be an edge consisting of the cycle; on the other hand, if an edge can be a part of a cycle, it must be a participant relation of the rule. Based on the above hypothesis, we define the proportion of edges existing in cycles to evaluate the rule-inferrable of KGs (i.e., the rule-inferrable indicator).

To verify our hypothesis, we conduct simulation experiments with a Family Tree KG~\cite{hohenecker2020ontology}, an artificially closed-world dataset generated with logical rules. By randomly selecting $N\%$ of triplets to replace with randomly sampled triplets, we evaluate their rule-inferrable indicators. As shown in Table~\ref{family-tree}, as the randomness increases, the proportion of edges appearing in cycles decreases and are all lower than in the standard Family Tree. These results indicate that the proportion of edges appearing in the rings can empirically measure the rule-inferrable of KGs.

\begin{table*}[h]
  \caption{Simulation results of family-tree datasets.}
  \label{family-tree}
  \begin{center}
\begin{small}

  \resizebox{0.65\columnwidth}{!}{
  \begin{tabular}{c|ccc}
    \toprule
    & \textbf{2-membered cycle} & \textbf{3-membered cycle} & \textbf{ $\leq$ 3-membered cycle} \\
\midrule
standard Family Tree  & 0.941  & 0.996  & 1.000  \\
random5\%    &0.850   & 0.958  & 0.960  \\
random10\%    &0.766   & 0.931  & 0.934  \\
random15\%    &0.684    &0.912  &0.915 \\

random20\%    & 0.611  &0.898  & 0.901  \\
random25\%    &0.542   &0.895  &0.898\\
random30\%    &0.479  &0.887   &0.891\\

 \bottomrule

  \end{tabular}}
      
\end{small}
\end{center}
\end{table*}

Next, we analyze the rule-inferrable on all datasets, i.e., FB15k-237, WN18RR, YAGO3-10, UMLS, Kinship and Family. The results are included in Table~\ref{rule-sensi}. We observe that: UMLS, Kinship and Family reach 100\% of 3-membered cycles while YAGO3-10 and WN18RR have a relatively low proportion, especially WN18RR, which is only about 17\%. Therefore, we can empirically conclude that compared to those KGs containing more general facts (FB15k-237, WN18RR and YAGO3-10), UMLS, Kinship and Family are more rule-inferrable datasets. Furthermore, the performance improvement of the RulE (\textit{emb with TransE.}) is more significant, which is consistent with the observation in our experiments (See Table~\ref{UK}).

\begin{table*}[h]
  \caption{The cycle proportion of edges on all datasets.}
  \label{rule-sensi}
  \begin{center}
\begin{small}

  \resizebox{0.65\columnwidth}{!}{
  \begin{tabular}{c|ccc}
    \toprule
     & \textbf{2-membered cycle} & \textbf{3-membered cycle} & \textbf{ $\leq$ 3-membered cycle} \\
\midrule
FB15k-237  & 0.344  & 0.856  & 0.877  \\
WN18RR    &0.389   & 0.177  & 0.452  \\
YAGO3-10    &0.569   & 0.179  & 0.698  \\

UMLS    & 0.676  &1.00  & 1.00  \\
Kinship    & 0.998   & 1.00  & 1.00  \\
Family &0.997 &0.954  &1.00 \\
 
 \bottomrule

  \end{tabular}}
      
\end{small}
\end{center}
\end{table*}

\section{Complexity analysis}\label{app:complexity}
This section analyzes the complexity of RulE. We use $d$ to denote hidden dimension and $\mathcal{E}$ is the set of relations (edges). 

During training, for the joint entity/relation/rule embedding stage, the amortized time of a single triplet or a logical rule is $O(d)$ due to linear operations. For the soft reasoning part, considering a query $( h, r, ?)$, RulE performs a BFS search from h to find all candidates and compute their grounding rule scores. We group triplets with the same $h$, $r$ together, where each group contains $|\mathcal{V}|$. For each group, we only need to use an MLP to get predictions, which takes $O(|\mathcal{E}|d^2)$ time. Thus, the amortized time for a single triplet is $O(\frac{|\mathcal{E}|d^2}{|\mathcal{V}|})$. 

During inference, we compute the final score with a weighted sum of the KGE score and the grounding rule score. Thus each triplets takes $O(\frac{|\mathcal{E}|d^2}{|\mathcal{V}|} + d)$ time.

The inference time of RulE and RNNLogic+ on different datasets is presented in Table~\ref{tab:complex}. We can see that RulE has similar inference time to RNNLogic+.

\begin{table*}[h]
  \caption{Inference time (in minutes) of RulE and RNNLogic+ on all datasets.}
  \label{tab:complex}
  \begin{center}
\begin{small}

  \resizebox{0.7\columnwidth}{!}{
  \begin{tabular}{c|cccccc}
    \toprule
    \textbf{Inference time} & \textbf{FB15k-237}  & \textbf{WN18RR} &\textbf{YAGO3-10} &  \textbf{UMLS}  & \textbf{Kinship}  & \textbf{Family} \\
\midrule
      RulE      &3.70    &3.10  &4.50 & 0.50 &0.75 &0.60 \\
     RNNLogic+ &4.10   &3.25 &4.88  &0.70 &0.90  &1.13\\

 \bottomrule

  \end{tabular}}
      
\end{small}
\end{center}
\end{table*}





      

\section{A variant of RulE with position-aware sum}\label{app:order}
In this section, considering the relation order of rules, we design a variant of RulE using position-aware sum and evaluate the variant based on TransE and RotatE.

It is obvious that 2D rotations and translations are commutative---they cannot model the non-commutative property of composition rules, which is crucial for correctly expressing the relation order of a rule. Take $\text{sister\_of}(x,y) \land \text{mother\_of}(y,z) \rightarrow \text{aunt\_of}(x,z)$ as an example. If we permute the relations in rule body, e.g., change ($\text{sister\_of} \land \text{mother\_of}$) to ($\text{mother\_of} \land \text{sister\_of}$), the rule is no longer correct. However, the above model will output the same score since $(\bm r_1 \circ \bm r_2) = (\bm r_2 \circ \bm r_1)$ and $(\bm r_1 + \bm r_2) = (\bm r_2 + \bm r_1)$. 

Therefore, to respect the relation order of logical rules, we use position-aware sum to model the relationship between logical rules and relations. Recall that $ \bm r \in \mathbb{C}^k$ is the embedding of relation and $g(\bm r)$ is to return the angle vector of relation $\bm r$. For each logical rule $\text R \colon \text r_1 \land \text r_2 \land \ldots \land \text r_l \rightarrow \text r_{l+1} $, we associate it with a rule embedding $\bm R = [\bm R^1, \bm R^2,..., \bm R^l], \bm R \in \mathbb{C}^{kl} $, where $l$ is the length of the logical rule and $[\cdot,\cdot]$ is concatenation operation. Based on the above definitions, we can formulate the distance function as:
\begin{equation}
\begin{aligned}
d(\bm r_1,\bm r_2,\ldots,\bm r_{l+1},\bm R) &= \parallel \sum_{j=1}^{l}\left(g(\bm r_k) \cdot g(\bm R^k)\right) \\ &- g(\bm r_{l+1}) \parallel,
\end{aligned}    
\end{equation}
where $\cdot$ is an element-wise product. Then we use Equation~(\ref{eq:rule-loss}) to further define the loss function of logical rules. 


Experimental results with TransE and RotatE are displayed in Table~\ref{tab:order}. RulE (\textit{emb\_o.}) is the new version that uses position-aware sum. From the results, we can see that RulE (\textit{emb\_o.}) almost obtains superior performance to the corresponding KGE models, again empirically demonstrating that jointly representing entity, relation and rule embeddings can improve the generalization of KGE. Moreover, the performance of RulE (\textit{emb\_o.}) is comparable with RulE (\textit{emb.}) in FB15k-237 and WN18RR. It also increases a lot in UMLS and Kinship, especially Kinship, which outperforms RulE (\textit{emb with TransE.}) with a 2.9\% improvement in MRR. The reason is probably that relation order plays an important role in modeling logical rules for rule-inferrable datasets (e.g., UMLS and Kinship).

\begin{table*}[h]
  \caption{Results of reasoning on FB15k-237, WN18RR, UMLS and Kinship. H@k is in \%.}
  \label{tab:order}
  \begin{center}
\begin{small}
  \resizebox{1.0\textwidth}{!}{
  \begin{tabular}{c|cccc|cccc|cccc|cccc}
    \toprule
    \multicolumn{1}{c}{\multirow{2}{*}{}} & \multicolumn{4}{|c|}{\textbf{FB15k-237}} & \multicolumn{4}{|c}{\textbf{WN18RR}} & \multicolumn{4}{|c}{\textbf{UMLS}} & \multicolumn{4}{|c}{\textbf{Kinship}} \\

  &MRR &H@1 &H@3 &H@10 &MRR &H@1 &H@3 &H@10  &MRR &H@1 &H@3 &H@10 &MRR &H@1 &H@3 &H@10\\

\midrule
 TransE   &0.329  &23.0  &36.9  &52.8  &0.222 &1.2 &39.9 &53.0 & 0.704  & 55.4 &82.6 &92.9  &0.300 &14.3 &35.2  &63.7  \\
 RulE (emb with TransE.)       & 0.346  &25.1  &38.5  &53.4  &0.242  &6.7 &37.8 &52.6    &0.748  &61.8   &85.1  &93.4 &0.347 &20.7  &39.8  &62.3    \\
 RulE (emb\_o with TransE.)    &0.336 &24.2 &37.2 &52.2  &0.220 &3.3  &37.2  &50.9 &0.765  &66.9  &82.9  &92.4  &0.376  &22.7  &42.4 &70.0    \\
 
 \midrule

 RotatE &0.337  &23.9  &37.4  &53.2  & 0.476  &43.1  &49.2 
 &56.2  &0.802 &69.6 &89.0 & 96.3   &0.672 &53.8 &76.4 &93.5  \\

 RulE (emb with RotatE.)   &0.337 &24.0 &37.5 &52.9 & 0.484  &44.3  & 49.9  & 56.3   & 0.807  & 70.6  &89.2  & 96.3  &0.675 &53.8  &77.1  & 93.7    \\
 RulE (emb\_o with RotatE.)  &0.338 &24.1 &37.6 &53.3  & 0.484  &44.1  &50.0 &56.7  & 0.809  &71.6  &88.3  &96.2  &0.676  &53.8  &77.2  &93.9    \\
 \bottomrule

  \end{tabular}}
      
\end{small}
\end{center}
\end{table*}

\begin{table*}[h]
\caption{Statistics of six datasets.}
\label{datasets}
\begin{center}
\resizebox{0.8\textwidth}{!}{
\begin{tabular}{c|ccccccc}
    \toprule
    Dataset    & \#Entities    & \#Relations    & \#Train   & \#Validation & \#Test & \#Rules  &\# length of rules  \\ 
    \midrule
    FB15k-237    & 14,541  & 237 & 272,115  & 17,535 & 20,466 &131,883 &$\leq$3\\
    WN18RR    & 40,943  & 11 & 86,835  & 3,034 & 3,134 &7,386 &$\leq$5\\
    YAGO3-10  &123,182 &37  &1,079,040 &5,000 &5,000 &7,351 &$\leq$2\\
    UMLS    & 135  & 46 & 1,959  & 1,306 & 3,264  &18,400 &$\leq$3\\
    Kinship    & 104  & 25 & 3,206  & 2,137 & 5,343 &10,000  &$\leq$3\\
    Family    &3007  & 12  &23,483  &2,038   &2,835  &2,400 &$\leq$3 \\
    \bottomrule
\end{tabular}}
\end{center}
\end{table*}

\begin{table*}[h]
\caption{Hyperparameter configurations of RulE on different datasets. }
\label{tab:params}
\begin{center}
\resizebox{0.75\textwidth}{!}{
\begin{tabular}{cc|cccccc}
    \toprule
    & \textbf{Hyperparameter}    & \textbf{FB15k-237}   &\textbf{WN18RR }   & \textbf{YAGO3-10}   & \textbf{UMLS} &\textbf{ Kinship} &\textbf{Family}   \\ 
    \midrule
     \multirow{10}{*}{\textbf{\shortstack{Joint\\embedding}}} &
    $k$    & 1000  & 500 & 500  & 2000 & 2000 &2000 \\
    & $bt$    &1024  & 512  &1024  &256  &256 &256 \\
    & $br$  &128   &256 &256 &256 &256 &256 \\
    & $\gamma_t$     &9  &6 &24  &6  &6   &6 \\
    & $\gamma_r$   &9  & 2 &24   &8  &5 &1  \\
    & $lr$    & 0.00005 &0.00005 &0.005  &0.0001  &0.0001 &0.0001 \\
    & $adv$  &1.0  &0.5  &1.0  &0.25  &0.25  &1.0\\
    & $\lambda$  &0  &0.1  &0  &0 &0.1  &1.0\\
    & $\alpha$ &3 &0.5  &10  &1 &3.0 & 1.0\\
    
    \midrule
    \multirow{3}{*}{\textbf{\shortstack{Soft rule\\reasoning}}} 
    & $lr$ &0.005 &0.005  &0.01  &0.0001  &0.0005 &0.0001\\
    & $gb$ &32 &32   &16    &16    &32  &32\\
    & $\beta$ &0.50  &0.60  &0.10  &0.20  &0.35 &0.35\\
    \bottomrule
\end{tabular}}
\end{center}
\end{table*}

\section{Different representations of entity-relation loss and relation-rule loss}\label{app:representations}
The entity-relation loss is defined in terms of the Hadamard product, while the relation-rule loss is defined in terms of $g( r)$. Essentially, the two representations are equivalent. We utilize distinct representations for the sake of convenience and to maintain consistency with the model's implementation. Following the RotatE~\cite{sun2019rotate} paper, the entity-relation loss (i.e., $t \approx h \circ r$) is defined in terms of the Hadamard product, which is equivalent to rotating the entity-vector with a relation-angle in 2D complex space. For relation-rule loss, if a logical rule $\text R:r_1 \land r_2 \land ... \land r_l \rightarrow r_{l+1}$ holds, we expect that $r_{l+1} \approx (r_1 \circ r_2 \circ ... \circ r_l ) \circ R$ . As RotatE restricts the modulus of each $r$'s dimension to be 1, the multiple rotations in the complex plane are equivalent to the summation of the corresponding angles (with the modulus unchanged), making it convenient to use the summation of angles in implementation. Therefore, we do not maintain modulus for $r$ and $R$ (since they are all 1) in our implementation, but only maintain their angular vectors, denoted by $g(r)$ and $g(R)$. To keep consistency with our implementation, it is beneficial to define the function $g(r)$ as the angle vector of relation $r$ and directly formulate the distance function in terms of angle vectors.

\section{Experiment setup}\label{statis}

\subsection{Data statistics}

The detailed statistics of six datasets for evaluation are provided in Table~\ref{datasets}. 
FB15k-237~\cite{toutanova2015observed}, WN18RR~\cite{dettmers2018convolutional} and YAGO3-10 are subsets of three large-scale knowledge graphs, FreeBase~\cite{Bollacker2008FreebaseAC} and WordNet~\cite{miller1995wordnet} and YAGO3~\cite{mahdisoltani2014yago3}. UMLS, Kinship and Family~\cite{kok2007statistical} are three benchmark datasets for statistical relational learning. For FB15k-237, WN18RR and YAGO3-10, we use the standard split. For Kinship and UMLS, we follow the data split from RNNLogic~\cite{qu2020rnnlogic} (i.e., split the dataset into train/validation/test with a ratio $3:2:5$) and report the results of some baselines taken from RNNLogic. For Family, we follow the split used by DRUM~\cite{sadeghian2019drum}. To ensure a fair comparison, we use RNNLogic to mine logical rules and rerun the reasoning predictor of RNNLogic+ with the same logical rules. Here, we consider chain rules, covering common logical rules in KG such as symmetry, composition, hierarchy rules, etc. Because inverse relations are required to apply rules, we preprocess the KGs to add inverse links. More introduction is included in Appendix~\ref{app:process}.

\subsection{Data process}\label{app:process}

 Most rules mined by rule mining systems are not chain rules. They usually need to be transformed into chain rules by inversing some relations. Considering $\text r_1(x,y) \land \text r_2(x,z) \rightarrow \text r_3(y,z)$ as an example, with replacing $\text r_1(x,y)$ with $\text r_1^{-1}(y,x)$, the rule can be converted into chain rule $\text r_1(y,x)^{-1} \land \text r_2(x,z) \rightarrow \text r_3(y,z)$. Based on the above, for data processing, we need to 
 add a inverse version triplet $(\text t, \text r^{-1}, \text h)$ for each triplet $(\text h, \text r, \text t)$, representing the inverse relationship $\text r^{-1}$ between entity $\text t$ and entity $\text h$.


\subsection{Evaluation protocol}\label{app:protocol}
During evaluation, for each test triplet $(h,r,t)$, we build two queries $(h,r,?)$ and $(t,r^{-1},?)$ with answer $t$ and $h$. For each query, we compute the KGE score and grounding rule score (Equation~\ref{eq:grounding}) for each candidate entity. As KGE scores and rule scores are scattered over different value ranges, we need to normalize the score before we compute the aggregated score. We map the grounding rule score to $[min, max]$ such that $min$ and $max$ are the minimum and maximum of KGE scores, i.e., map to the range of KGE scores. Then RulE weighted sums over both scores (i.e., $\beta \ast s_g + (1-\beta)\ast s_{t_{norm}} $). Once we have the final score for all candidate answers, consider the situation that many entities might be assigned the same score. Following RNNLogic~\cite{qu2020rnnlogic}, we first random shuffles of those entities which receive the same score and then compute the expectation of evaluation metric over them.

\begin{figure}[t]
\centering
\subfloat[FB15k-237]{\label{fig:beta1}\includegraphics[width=0.4\columnwidth]{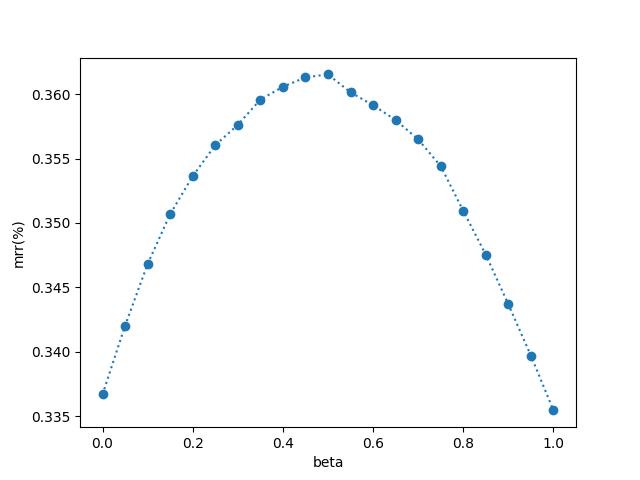}}
\subfloat[WN18RR]{\label{fig:beta2}\includegraphics[width=0.4\columnwidth]{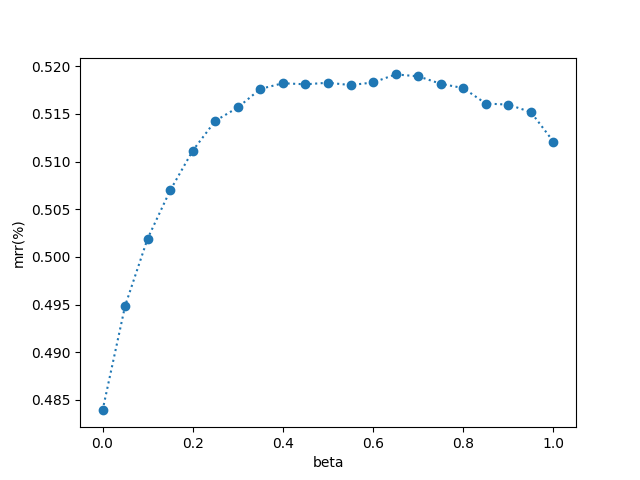}}

\caption{(a) and (b) show the MRR results of RulE with varying $\beta$ on FB15k-237 and WN18RR.}
\label{Fig:beta}

\end{figure}

\begin{table}[h]
    \centering
    \resizebox{0.9\linewidth}{!}{
    \begin{tabular}{c|c|c|c|c|c}
    \toprule
     MRR &FB15k-237 &WN18RR &UMLS &Kinship &family \\  
     \midrule
     NBFNet &0.415 &0.551 &0.922 &0.635 &0.990  \\
     RulE &0.362 &0.519 &0.867 &0.736 &0.984  \\
     \bottomrule
    \end{tabular}}
    \caption{Comparison NBFNet with RulE.}
    \label{tab:nbfnet}
\end{table}

\subsection{Hyperparameter optimization}\label{app:hyper}
We search for parameters according to validation set performance. For above baselines, we carefully tune the parameters and achieve better results than reported in RNNLogic~\cite{qu2020rnnlogic}. To ensure a fair comparison, in the KGE part of RulE, we use the same parameters as those used in TransE and RotatE without further tuning them. When comparing RulE (\textit{rule.}) with RNNLogic+ (\textit{w/o emb.}), we use the same logical rules mined from RNNLogic~\cite{qu2020rnnlogic}. Note that the reported results for TransE and RotatE are indeed based on their best parameter settings, where we carefully tuned their parameters such that our reported results for TransE and RotatE are even higher than those reported in RNNLogic~\cite{qu2020rnnlogic}. However, in the KGE part of RulE, we use the same parameters as those used in TransE and RotatE without further tuning them. So the truth is, we did not adopt TransE/RotatE settings tuned on RulE for TransE/RotatE, but on the contrary, adopt TransE/RotatE settings tuned on themselves for RulE. This should bring disadvantages to RulE, yet we still observe improved performance.

The hyperparameters are tuned by the grid search, The range is set as follows: embedding dimension $k \in \{500,1000,2000\}$, batch size of triplets and rules $b \in \{256, 512, 1024\}$, the weight balancing two losses ($L_t$ and $L_r$) $\alpha \in \{0.5, 1, 2, 3, 4, 5\}$, triplet margin and rule margin $\gamma_t, \gamma_r \in [0:30:1]$ and the weight balancing embedding-based and rule-based reasoning $\beta \in [0:0.05:1] $. The optimal parameter configurations for different datasets for RulE (\textit{emb \& rule.}) can be found in Table~\ref{tab:params}, including embedding dimension $k$, batch size of triplets $bt$, batch size of rules $br$, fix margin of triplets $\gamma_t$, fix margin of triplets $\gamma_r$, learning rate $lr$, self-adversarial sampling temperature $adv$, regularization coefficient $\lambda$, the weight balancing the importance of rules in joint loss function (Equation~\ref{lossall}) $\alpha$, batch size in soft rule reasoning $gb$ and the weight of inference process (Equation~\ref{equ:final}) $\beta$. Note that we use RotatE as the KGE model. 

\section{Experiment details}

\subsection{Embedding logical rules helps KGE}\label{abl:transe}
This section discusses the effectiveness of rule embedding on KGE.
As shown in Table~\ref{apptable:emb}, the two variants using TransE and ComplEx as KGE models are denoted by RulE (\textit{emb with TransE.}) and RulE (\textit{emb with ComplEx.}), respectively. 
 They both obtain superior performance to the corresponding KGE models. 
 
 We also further compare with other rule-enhance KGE models. In the experiment setup, RulE (\textit{emb with TransE.}) uses the same logical rules as KALE~\cite{guo2016jointly}; RulE (\textit{emb with ComplEx.}) uses the same logical rules as ComplEx-NNE-AER~\cite{ding2018improving}. The comparison shows that RulE (\textit{emb with TransE.}) yields more accurate results than KALE. For RulE (\textit{emb with ComplEx.}), although it does not outperform ComplEx-NNE+AER (probably because it additional injects the regularization terms on entities but RulE does not), compared to RUGE, RulE (\textit{emb with ComplEx.}) also obtains 2\% improvement in MRR on FB15k as well as comparable results on WN18. 


For a fair comparison, RulE (\textit{emb. TransE}) applies the same logical rules as KALE; RulE (\textit{emb. ComplEx}) uses the same logical rules as ComplEx-NNE-AER. 

\begin{table*}[h]
  \caption{Results of reasoning on FB15k and WN18. H@k is in \%. [*] means the numbers are taken from~\cite{guo2018knowledge} and \cite{ding2018improving}. [$^\dagger$] means we rerun the methods with the same evaluation process.}
  \label{apptable:emb}
  \begin{center}
\begin{small}

  \resizebox{0.7\textwidth}{!}{
  \begin{tabular}{c|cccc|cccc}
    \toprule
    \multicolumn{1}{c}{\multirow{1}{*}} & \multicolumn{4}{|c|}{\textbf{FB15k}} & \multicolumn{4}{|c}{\textbf{WN18}} \\

  &MRR &H@1 &H@3 &H@10 &MRR &H@1 &H@3 &H@10 \\

\midrule
 $\text{TransE}^\dagger$ &0.730  &64.6  &79.2  &86.4 &0.772  &\textbf{70.5}  &80.8  &92.2   \\
 $\text{KALE}^*$   &0.523 &38.3 &61.6 &76.2 &0.662  &- &85.5  &93.0   \\
 RulE (\textit{emb with TransE.}) &\textbf{0.734} &\textbf{65.0} &\textbf{79.9} &\textbf{86.9}  &\textbf{0.775} &67.2 &\textbf{86.2} &\textbf{95.0}\\
 \midrule
 $\text{ComplEx}^\dagger$ &0.766  &69.7  &81.3  &88.3  &0.898  &85.4  &92.6  &\textbf{95.2} \\

  $\text{RUGE}^*$   &0.768  &70.3  &81.5  &86.5   &0.943  &-  &-  &94.4\\
$\text{ComplEx-NNE+AER}^*$ &\textbf{0.803}  &\textbf{76.1}  & 83.1 &87.4 &\textbf{0.943} &\textbf{94.0} &\textbf{94.5} &94.8  \\
 
 RulE (\textit{emb with ComplEx.}) &0.788 &72.4  &\textbf{83.3}  &\textbf{89.6} &0.928  &91.9  &93.5  &94.4 \\
 \bottomrule

  \end{tabular}}
      
\end{small}
\end{center}
\end{table*}
\begin{table*}[h]
  \caption{Ablation results on FB15k-23 and WN18RR datasets. H@k is in \%.}
  \begin{center}\label{ablation-res1}
\begin{small}

  \resizebox{0.7\textwidth}{!}{
  \begin{tabular}{c|cccc|cccc}
    \toprule
    \multicolumn{1}{c}{\multirow{2}{*}{}} & \multicolumn{4}{|c|}{\textbf{FB15k-237}} & \multicolumn{4}{|c}{\textbf{WN18RR}}  \\

  &MRR &H@1 &H@3 &H@10 &MRR &H@1 &H@3 &H@10  \\

  \midrule
 standard  &0.335 &24.9  &36.9 &50.4  &{0.514} &{47.3} &{53.3} &{59.7}  \\
sum (w/o MLP)  &0.276  &19.8  &30.2  &42.9  &0.390  &32.7  &41.9  &50.9    \\

max (w/o MLP)  &0.256  &18.4  &27.7  &39.7  & 0.294  &23.4 &31.5 &41.4   \\
 hard-encoding  &0.330  &24.3   &36.3  &50.2   &0.496 &45.4 &51.5  &57.7     \\
 \bottomrule

   \end{tabular}}
      
\end{small}
\end{center}
\end{table*}

\begin{table*}[!h]
  \caption{Ablation results on UMLS, Kinship and Family datasets. H@k is in \%.}
  \begin{center}\label{ablation-res2}
\begin{small}

  \resizebox{0.8\textwidth}{!}{
  \begin{tabular}{c|cccc|cccc|cccc}
    \toprule
    \multicolumn{1}{c}{\multirow{2}{*}{}} &  \multicolumn{4}{|c}{\textbf{UMLS}} & \multicolumn{4}{|c}{\textbf{Kinship}} & \multicolumn{4}{|c}{\textbf{Family}} \\

  &MRR &H@1 &H@3 &H@10 &MRR &H@1 &H@3 &H@10  &MRR &H@1 &H@3 &H@10 \\

  \midrule
 standard  &0.827 &74.9 &88.9 &95.5  &0.673 &52.8 &77.5 &95.0 &0.975 &96.7  &98.5  &98.6 \\

 sum (w/o MLP)  &0.587 &46.1 &65.7 &82.0 &0.591  &44.3  &67.4  &90.0 &0.877  &81.2 &92.9 &97.6   \\
 max (w/o MLP)  &0.346  &23.1  &36.4 &58.7 &0.372  &21.8  &40.7  &74.7 &0.748  &63.9 &82.7  &94.9  \\
 hard-encoding   &0.791  &69.5  &86.7  &94.6   &0.643  &49.1  &74.5 &94.0   &0.973  &96.2 &98.4 &98.6 \\
 \bottomrule

  \end{tabular}}
      
\end{small}
\end{center}
\end{table*}

 \subsection{Sensitivity analysis of beta}\label{app:beta}
To analyze how the hyperparameter $\beta$ balances the weights of embedding-based and rule-based reasoning (defined in Equation~(\ref{equ:final})), we conduct experiments for RulE under varying $\beta$. Figure~\ref{fig:beta1} and \ref{fig:beta2} show the results on Fb15k-237 and WN18RR.

With the increase of $\beta$, the performance of RulE first improves and then drops on both datasets. This is because the information captured by logical rules and knowledge graph embedding is complementary, thus combining embedding-based and rule-based methods can enhance knowledge graph reasoning. Moreover, the trend of $\beta$ for the performance on the two datasets is different (FB15k-237 tends to drop faster than WN18RR). We think that in WN18RR, information captured by the rule-based method may be more than embedding-based, leading that the rule-based method is more predominant in WN18RR ($\beta = 0.6$).

\subsection{More results of ablation study}\label{appendix:ablation}
More results of ablation study are presented in Table~\ref{ablation-res1} and \ref{ablation-res2}.

\subsection{Comparison NBFNet with RulE}\label{app:nbfnet}

We follow the results of FB15k-237 and WN18RR reported in NBFNet and conduct additional experiments on UMLS, Kinship and family datasets. The results (MRR) are shown in Table~\ref{tab:nbfnet}:

NFBNet has better results than RulE on FB15k-237, WN18RR and UMLS. However, RulE achieves comparable or higher performance than NBFNet on Kinship and family, especially on Kinship, where RulE obtains about 10\% absolute MRR gain. This might be explained by that Kinship and family contain more rule-inferrable facts while WN18RR and FB15k-237 consist of more general facts (a more detailed discussion is given in Appendix~\ref{cycle}). This indicates that our method RulE is more favorable for knowledge graphs where rules play an important role, which is expected as it leverages rules explicitly. Another advantage of RulE is the ability to use prior/domain knowledge, while GNN-based methods cannot leverage prior/domain knowledge presented as logical rules. Moreover, RulE is more interpretable on rule-level than GNN methods, which is still valuable in certain domains. Although NBFNet is also interpretable, RulE's interpretability is on rule level while that of NBFNet is on path level. For example, when the KG system desires high interpretability (such as those in medical applications), each inferred knowledge must be accompanied with which exact rules are responsible for the inference, otherwise the doctors are hard to trust it. In contrast, GNN methods (such as NBFNet) are only interpretable on path-level instead of rule-level. Take "Alice is Bob's mother" as an example, GNN methods might tell us the path "Alice is David's mother" and "David is Bob's brother" is activated during the inference, while our RulE can not only tell us that this path is activated, but also the rule $\forall x,y,z: \text{mother}(x,y) \land \text{brother}(y,z) \rightarrow \text{mother}(x,z)$ is responsible behind the prediction. 

In summary, although NBFNet demonstrates state-of-the-art performance on many KGs, we still believe a hybrid method that can explicitly model and leverage logical rules is desired and worth studying.


\section{Theoretical analysis and case studies}\label{case studies}

As mentioned in the main body, the rule embeddings are not only used to regularize the embedding learning. On the other hand, with the rule embeddings, RulE can compute the confidence score for each logic rule, which enhances the original hard rule-based reasoning process through soft rule confidence. Additionally, combining the jointly trained KGE and the confidence-enhanced rule-based reasoning, we arrive at a final neural-symbolic model achieving superior performance on many datasets.

Consider the rule $r_1(x,y) \land r_2 (y,z)  \rightarrow r_3(x,z)$ as an example, where $x$, $y$, $z$ represent specific entities. Given three facts, we obtain $y=x \circ r_1$; $z=y\circ r_2$; $z=x\circ r_3$. Combining these equations, we deduce $r_1 \circ r_2 =r_3$.  However, those mined rules may not be confidently correct. Thus, we assign a residual embedding as a rule embedding to each logical rule, i.e., $r_1 \circ r_2 \circ R =r_3$. By adding additional constraints that relations should satisfy, rule loss provides a regularization to the triplet (KGE) loss, improving the generalization of KGE. Meanwhile, with the relation and rule embeddings, RulE can further give a confidence score to each rule, which reflects how consistent a rule is with the existing facts and enables performing the rule inference process in a soft way. This provides an explanation of why RulE is better than naive combination methods.

We further provide some case studies illustrating the confidence scores of logical rules learned by RulE on the family dataset.
\newline
\newline
$
(1)\ \text{brother}(x,y) \land \text{brother}(z,y)\land \text{mother}(t,z) \\  ~~~~~~~~~~~~~~~~~~~~~~~~~~~~~~~~~~~~~~~~~~~~~~~~\rightarrow \text{son}(x,t)~~~~~0.932 \\ 
$
$
(2)\ \text{brother}(y,x) \land \text{brother}(y,z)\land \text{father}(t,z)
\\
~~~~~~~~~~~~~~~~~~~~~~~~~~~~~~~~~~~~~~~~~~~~~~~~\rightarrow \text{son}(x,t)~~~~~0.798 \\ 
$
$
(3)\ \text{mother}(x,y) \land \text{brother}(z,y)\\ ~~~~~~~~~~~~~~~~~~~~~~~~~~~~~~~~~~~~~~~~\rightarrow \text{mother}(x,z)~~~~~0.834 \\ 
$
$
(4)\ \text{wife}(x,y) \land \text{son}(z,y) \rightarrow \text{mother}(x,z) \ \ \ \ \ \ 0.589 \\
$

Ideally, rules with higher success probability should yield higher confidence scores. For instance, rule (1) has a higher confidence score than rule (2) because the $x$ in rule (2) could also be the daughter of $t$, while the $x$ in rule (1) must be male because $x$ is $y$'s brother. Our RulE successfully learns them out. Another example is rule (3) and rule (4). They both infer $x$ is $z$'s mother, but rule (4) is less confident because $x$ can also be $z$'s stepmother.

\end{document}

%% file: config.tex
\usepackage{enumitem}
\usepackage{subfig}
\usepackage{graphicx} 
\usepackage{booktabs} 
\usepackage{subfloat}
\usepackage{multirow}
\usepackage{floatrow}
\usepackage{extarrows}
\usepackage{acronym}
\usepackage{xspace}
\usepackage{cleveref}
\usepackage[tableposition=top]{caption}
\newfloatcommand{capbtabbox}{table}[][\FBwidth]
\usepackage{titletoc}
\usepackage{appendix}

\makeatletter
\DeclareRobustCommand\onedot{\futurelet\@let@token\@onedot}
\def\@onedot{\ifx\@let@token.\else.\null\fi\xspace}

\makeatother

\acrodef{llm}[LLM]{Large Language Model}
\acrodef{kg}[KG]{knowledge graph}

\DeclareMathOperator{\LN}{LN}
\DeclareMathOperator{\AGG}{AGG}
\DeclareMathOperator{\MLP}{MLP}

\newcommand{\answerTODO}[1][]{\textcolor{red}{\bf [TODO]}}

%% file: math_commands.tex
\usepackage{amsmath,amsfonts,bm}

\def\eqref#1{equation~\ref{#1}}

\def\1{\bm{1}}

\DeclareMathAlphabet{\mathsfit}{\encodingdefault}{\sfdefault}{m}{sl}
\SetMathAlphabet{\mathsfit}{bold}{\encodingdefault}{\sfdefault}{bx}{n}

